\documentclass[letterpaper, 10 pt, conference]{ieeeconf}  

\IEEEoverridecommandlockouts                              
\usepackage{enumerate}
\usepackage{amsmath} 
\usepackage{amssymb} 
\usepackage{amsfonts}

\usepackage{amsthm}

\usepackage{graphicx}
\usepackage{graphics} 
\usepackage{epsfig} 
\usepackage{accents}
\usepackage{algorithm}
\usepackage{algpseudocode}
\usepackage{subcaption}
\usepackage{array}

\usepackage{color}
\makeatletter
\let\NAT@parse\undefined
\makeatother

\usepackage{cite}

\usepackage{nohyperref}
\usepackage{url}
\usepackage{comment}

	\newtheoremstyle{myplain}
	  {}
	  {}
	  {\itshape}
	  {}
	  {\bfseries}
	  {}
	  {5pt plus 1pt minus 1pt}
	  {}

	\newtheoremstyle{mydefinition}
	  {}
	  {}
	  {\normalfont}
	  {}
	  {\bfseries}
	  {}
	  {5pt plus 1pt minus 1pt}
	  {}
  
	\theoremstyle{myplain}
    \newtheorem{assumption}{Assumption}

	\newtheorem{theorem}{Theorem}
	
	\newtheorem{proposition}{Proposition}

	\theoremstyle{mydefinition}
	\newtheorem{definition}{Definition}

\algdef{SE}[DOWHILE]{Do}{doWhile}{\algorithmicdo}[1]{\algorithmicwhile\ #1}%

\include{notation}
\newcommand{\robotradius}{r} 
\newcommand{\sensorrange}{R} 

\newcommand{\robotposition}{\mathbf{x}} 
\newcommand{\robotpositionmodel}{\mathbf{y}} 

\newcommand{\robotpositionunicycle}{\overline{\mathbf{x}}} 
\newcommand{\robotorientation}{\psi} 
\newcommand{\robotpositionunicyclemodel}{\overline{\mathbf{y}}} 
\newcommand{\robotorientationmodel}{\varphi} 

\newcommand{\goalposition}{\mathbf{x}_d} 
\newcommand{\goalpositiondot}{\dot{\mathbf{x}}_d} 
\newcommand{\goalpositionmodel}{\mathbf{y}_d}
\newcommand{\goalpositionmodeldot}{\dot{\mathbf{y}}_d}

\newcommand{\enclosingworkspace}{\mathcal{W}_e} 
\newcommand{\workspace}{\mathcal{W}} 
\newcommand{\enclosingfreespace}{\mathcal{F}_e} 
\newcommand{\freespace}{\mathcal{F}} 
\newcommand{\freespacesemantic}{\mathcal{F}_{sem}^\hybridmode} 
\newcommand{\freespacemapped}{\mathcal{F}_{map}^\hybridmode} 
\newcommand{\freespacemappedhat}{\hat{\mathcal{F}}_{map}^\hybridmode} 
\newcommand{\freespacemodel}{\mathcal{F}_{model}^\hybridmode} 

\newcommand{\obstacle}{\tilde{O}} 
\newcommand{\obstacleset}{\tilde{\mathcal{O}}} 
\newcommand{\obstaclesetdilated}{\mathcal{O}} 

\newcommand{\knownobstacle}{\tilde{P}} 
\newcommand{\knownobstacleset}{\tilde{\mathcal{P}}} 
\newcommand{\knownobstaclesetindex}{\mathcal{N}_{\mathcal{P}}}
\newcommand{\knownobstaclesetphysical}{\knownobstacleset_{\hybridmode}} 
\newcommand{\knownobstaclesetdilatedmappedindex}{\mathcal{J}^\hybridmode}
\newcommand{\knownobstaclesetdilatedmappedintrusionindex}{\mathcal{J}_{\mathcal{B}}^\hybridmode}
\newcommand{\knownobstaclesetdilatedmappeddiskindex}{\mathcal{J}_{\mathcal{D}}^\hybridmode}
\newcommand{\knownobstacledilated}{P} 
\newcommand{\knownobstaclesetdilated}{\mathcal{P}} 
\newcommand{\knownobstaclecardinality}{N_P} 

\newcommand{\knownobstaclesetdilatedsemantic}{\mathcal{P}_{sem}^\hybridmode} 
\newcommand{\knownobstaclesetdilatedmapped}{\mathcal{P}_{map}^\hybridmode} 
\newcommand{\knownobstacledilatedmappedintrusion}{B} 
\newcommand{\knownobstaclesetdilatedmappedintrusion}{\mathcal{B}_{map}^\hybridmode} 
\newcommand{\knownobstaclesetdilatedmappeddisk}{\mathcal{D}_{map}^\hybridmode} 

\newcommand{\unknownobstacleset}{\tilde{\mathcal{C}}} 
\newcommand{\unknownobstaclesetindex}{\mathcal{N}_{\mathcal{C}}}
\newcommand{\unknownobstacledilated}{C} 
\newcommand{\unknownobstaclesetdilated}{\mathcal{C}} 
\newcommand{\unknownobstaclecardinality}{N_C} 

\newcommand{\unknownobstaclesetdilatedsemantic}{\mathcal{C}_{sem}} 
\newcommand{\unknownobstaclesetdilatedsemanticindex}{\mathcal{J}_{\mathcal{C}}}
\newcommand{\unknownobstaclesetdilatedmapped}{\mathcal{C}_{map}} 

\newcommand{\betaclearance}{\varepsilon}
\newcommand{\implicitgeneric}{\beta}

\newcommand{\diffeogeneric}{\mathbf{h}} 
\newcommand{\diffeo}{\mathbf{h}^{\hybridmode}} 
\newcommand{\diffeounicycle}{\overline{\diffeogeneric}^{\hybridmode}}
\newcommand{\diffeopurging}[1]{\mathbf{h}^{\hybridmode}_{#1}} 
\newcommand{\diffeocomposition}{\mathbf{g}^{\hybridmode}} 
\newcommand{\diffeoroot}{\hat{\mathbf{h}}^{\hybridmode}} 

\newcommand{\diffeocenter}[1]{\mathbf{x}_{#1}^*} 
\newcommand{\diffeoradius}[1]{\rho_{#1}} 

\newcommand{\polygontree}[1]{\mathcal{T}_{#1}} 
\newcommand{\polygonvertices}[1]{\mathcal{V}_{#1}} 
\newcommand{\polygonedges}[1]{\mathcal{E}_{#1}} 
\newcommand{\freespacemappedpurging}[1]{\mathcal{F}_{map,#1}^\hybridmode} 
\newcommand{\innerpolygon}[1]{\mathcal{Q}_{#1}} 
\newcommand{\outerpolygon}[1]{\overline{\mathcal{Q}}_{#1}} 
\newcommand{\innerpolygonimplicit}[1]{\gamma_{#1}} 
\newcommand{\outerpolygonimplicit}[1]{\delta_{#1}} 
\newcommand{\innerpolygonsigma}[1]{\sigma_{\innerpolygonimplicit{#1}}} 
\newcommand{\outerpolygonsigma}[1]{\sigma_{\outerpolygonimplicit{#1}}} 
\newcommand{\innerpolygontune}[1]{\mu_{\innerpolygonimplicit{#1}}}
\newcommand{\innerpolygondistance}[1]{\epsilon_{#1}}
\newcommand{\outerpolygontune}[1]{\mu_{\outerpolygonimplicit{#1}}}
\newcommand{\switch}[1]{\sigma_{#1}} 
\newcommand{\switchresidual}{\sigma_d}
\newcommand{\deformingfactor}[1]{\nu_{#1}} 
\newcommand{\sharednormal}[1]{\mathbf{n}_{#1}}

\newcommand{\controlfullyactuated}{\mathbf{u}} 
\newcommand{\controlunicycle}{\overline{\controlfullyactuated}} 
\newcommand{\controlfullyactuatedmodel}{\mathbf{v}} 
\newcommand{\controlunicyclemodel}{\overline{\controlfullyactuatedmodel}} 
\newcommand{\linearinput}{v} 
\newcommand{\angularinput}{\omega} 
\newcommand{\linearinputmodel}{\hat{\linearinput}} 
\newcommand{\angularinputmodel}{\hat{\angularinput}} 
\newcommand{\angletransform}{\xi^{\hybridmode}} 
\newcommand{\directionvector}{\mathbf{e}}
\newcommand{\projection}[2]{\mathrm{\Pi}_{#1}(#2)} 
\newcommand{\localfreespace}[1]{\mathcal{LF}(#1)} 

\newcommand{\hybridmode}{\mathcal{I}} 

\newcommand{\hybridfreespacemodemodel}{\mathcal{F}_{model}^\hybridmode} 

\newcommand{\ball}[2]{\mathsf{B}(#1,#2)}
\newcommand{\ballclosure}[2]{\overline{\mathsf{B}(#1,#2)}}

\title{\LARGE \bf
Technical Report: Reactive Semantic Planning in Unexplored Semantic Environments Using Deep Perceptual Feedback
}

\author{Vasileios Vasilopoulos, Georgios Pavlakos, Sean L. Bowman, J. Diego Caporale,\\ Kostas Daniilidis, George J. Pappas, Daniel E. Koditschek
\thanks{This work was supported by AFRL grant FA865015D1845 (subcontract 669737-1), AFOSR grant FA9550-19-1-0265 (Assured Autonomy in Contested Environments), and ONR grant \#N00014-16-1-2817, a Vannevar Bush Fellowship held by the last author, sponsored by the Basic Research Office of the Assistant Secretary of Defense for Research and Engineering. The authors thank Diedra Krieger for assistance with video recording.}%
\thanks{The authors are with the GRASP Lab, University of Pennsylvania, Philadelphia, PA 19104,
        {\tt\small \{vvasilo, pavlakos, seanbow, jdcap, kostas, pappasg, kod \}@seas.upenn.edu}}
}

\begin{document}

\maketitle
\thispagestyle{empty}
\pagestyle{empty}

\begin{abstract}
This paper presents a reactive planning system that enriches the topological representation of an environment with a tightly integrated semantic representation, achieved by incorporating and exploiting advances in deep perceptual learning and probabilistic semantic reasoning. Our architecture combines object detection with semantic SLAM, affording robust, reactive logical as well as geometric planning in unexplored environments. Moreover, by incorporating a human mesh estimation algorithm, our system is capable of reacting and responding in real time to semantically labeled human motions and gestures. New formal results allow tracking of suitably non-adversarial moving targets, while maintaining the same collision avoidance guarantees. We suggest the empirical utility of the proposed control architecture with a numerical study including comparisons with a state-of-the-art dynamic replanning algorithm, and physical implementation on both a wheeled and legged platform in different settings with both geometric and semantic goals.
\end{abstract}


\section{INTRODUCTION}
\label{sec:introduction}

\subsection{Motivation and Prior Work}

Navigation is a fundamentally topological problem \cite{farber_topology} reducible to purely reactive (i.e., closed loop state feedback based) solution, given perfect prior knowledge of the environment \cite{rimon1992}. For geometrically simple environments, ``doubly reactive'' methods that reconstruct the local obstacle field on the fly \cite{Paternain_Koditschek_Ribeiro_2017,Ilhan_Johnson_Koditschek_2018}, or operate with no need for such reconstruction at all \cite{Arslan_Koditschek_2018}, can guarantee collision free convergence to a designated goal with no need for further prior information. However, imperfectly known environments presenting densely cluttered or non-convex obstacles have heretofore required incremental versions of random sampling-based tree construction \cite{karaman_frazzoli_ICRA2012} whose probabilistic completeness can be slow to be realized in practice, especially when confronting settings with narrow passages \cite{noreen-khan-habib-2016}.

\begin{figure}[h!]
\captionsetup{width=\linewidth,font=footnotesize}
\centering
\includegraphics[width=1.0\columnwidth]{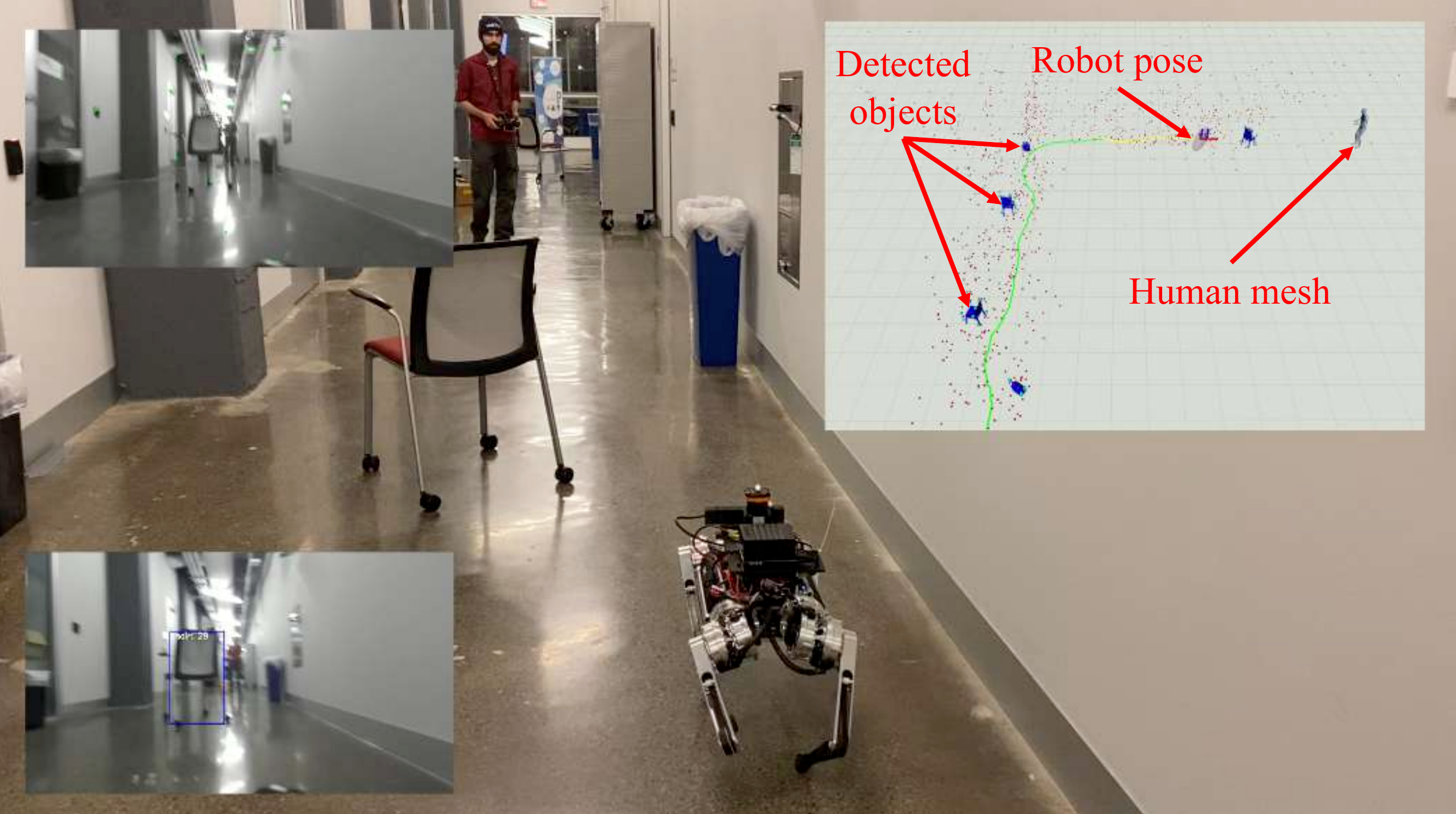}
\caption{Ghost Spirit \cite{ghostspirit} following a human, while avoiding some familiar and some novel obstacles in a previously unexplored environment.  Familiar obstacles are recognized and localized using visually detected semantic keypoints (bottom  left inset) \cite{Pavlakos2017}, combined with geometric features (top  left inset) \cite{Bowman2017} and avoided by a local deformation of space (Fig. \ref{fig:purging_transformation}) that brings them within the scope of a doubly reactive navigation algorithm \cite{Arslan_Koditschek_2018}.  Novel obstacles are detected by LIDAR and assumed to be convex, thus falling within the scope of \cite{Arslan_Koditschek_2018}.  Formal guarantees are summarized in Theorems \ref{theorem:control_fullyactuated} and \ref{theorem:control_se2} of Section \ref{sec:reactiveplanning}, and experimental settings are summarized in Fig. \ref{fig:environments}.}
\label{fig:experiment_spirit}
\vspace{-14pt}
\end{figure}

\begin{figure*}[t]
\captionsetup{width=\linewidth,font=footnotesize}
\centering
\includegraphics[width=1.0\textwidth]{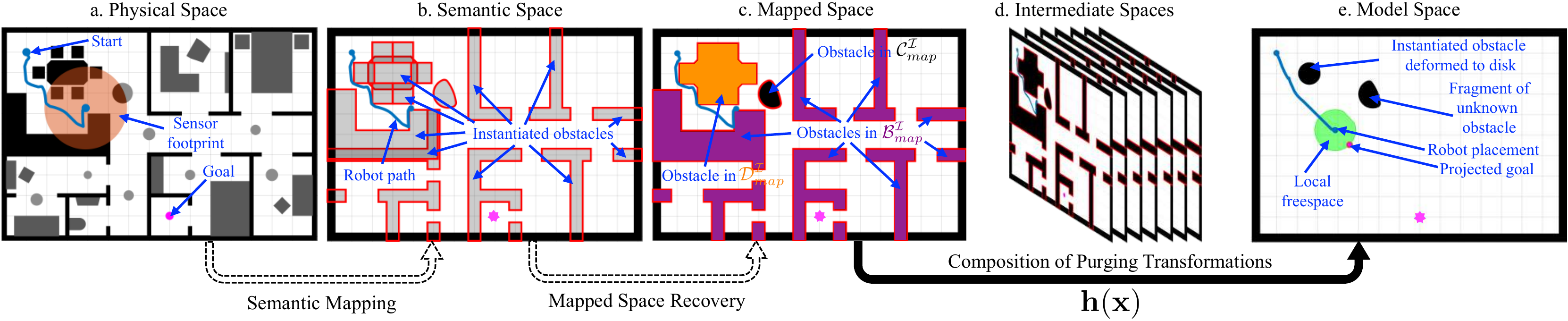}
\caption{Snapshot Illustration of Key Ideas, following \cite{vasilopoulos_pavlakos_schmeckpeper_daniilidis_koditschek_2019}: The robot moves in the physical space, in an environment with known exterior boundaries (walls), toward a goal (pink) discovering along the way (black) both familiar objects of known geometry but unknown location (dark grey) and unknown obstacles (light grey), with an onboard sensor of limited range (orange disk). As in \cite{vasilopoulos_pavlakos_schmeckpeper_daniilidis_koditschek_2019}, these obstacles are processed by the perceptual pipeline (Fig. \ref{fig:algorithm}) and stored permanently in the semantic space if they have familiar geometry, or temporarily, with just the corresponding sensed fragments, if they are unknown. The consolidated obstacles (formed by overlapping catalogued obstacles from the semantic space), along with the perceptually encountered components of the unknown obstacles, are again stored in the mapped space. A change of coordinates, $\diffeogeneric$, entailing an online computation greatly streamlined relative to its counterpart in \cite{vasilopoulos_pavlakos_schmeckpeper_daniilidis_koditschek_2019} deforms the mapped space to yield a geometrically simple but topologically equivalent model space. This new change of coordinates defines a vector field on the model space, which is transformed in realtime through the diffeomorphism to generate the input in the physical space.}
\label{fig:diffeo_idea}
\vspace{-14pt}
\end{figure*}

Monolithic end-to-end learning approaches to navigation -- whether supporting metric \cite{Gupta_2017} or topological \cite{savinov-dosovitskiy-koltun-2018} representations of the environment -- suffer from the familiar problems of overfitting to specific settings or conditions. More modular data driven methods that separate the recruitment of learned visual representation to support learned control policies achieve greater generalization \cite{shen-xu-zhu--guibas-feifei-savarese-2019}, but even carefully modularized approaches that handcraft the interaction of learned topological global plans with learned reactive motor control in a physically informed framework \cite{meng-ratliff-xiang-fox-2019} cannot bake into their architectures the exploitation of crucial properties that robustify design and afford guaranteed policies.

Unlike the problem of safe navigation in a completely known environment, the setting where the obstacles are not initially known and are incrementally revealed online has so far received little theoretical interest. Some few notable exceptions include considerations of optimality in unknown spaces \cite{janson-hu-pavone-2018}, online modifications to temporal logic specifications \cite{lahijanian-2016} or deep learning algorithms \cite{bajcsy2019efficient} that assure safety against obstacles, or the use of trajectory optimization along with offline computed reachable sets \cite{kousik-vaskov-bu-johnson-vasudevan-2018} for online policy adaptations. However, none of these advances (and, to the best of our knowledge, no work prior to \cite{vasilopoulos_pavlakos_schmeckpeper_daniilidis_koditschek_2019}) has achieved simultaneous guarantees of obstacle avoidance and convergence. In this paper, we extend these guarantees to the setting of (non-adversarial) moving targets, while also affording semantic specifications - capabilities that have not been heretofore available, even in settings with simple obstacles.

\subsection{Summary of Contributions}

\subsubsection{Architectural Contributions}
In \cite{vasilopoulos_pavlakos_schmeckpeper_daniilidis_koditschek_2019}, we introduced a Deep Vision based object recognition system \cite{Pavlakos2017} as an ``oracle'' for informing a doubly reactive motion planner \cite{Arslan_Koditschek_2018,vasilopoulos_koditschek_WAFR2018}, incorporating a Semantic SLAM engine \cite{Bowman2017} to integrate observations and semantic labels over time. Here, we extend this architecture in two different ways (see Fig. \ref{fig:algorithm}). First, new formal advances, replacing the computationally expensive triangulation on the fly \cite{vasilopoulos_pavlakos_schmeckpeper_daniilidis_koditschek_2019} with convex decompositions of obstacles as described below, streamline the reactive computation, enabling robust online and onboard implementation (perceptual updates at 4Hz; reactive planning updates at 30Hz), affording tight realtime integration of the Semantic SLAM engine. Second, we incorporate a separate deep neural net that captures a wire mesh representation of encountered humans \cite{kolotouros2019learning}, enabling our reactive module to track and respond in realtime to semantically labeled human motions and gestures.

\subsubsection{Theoretical Contributions}

We introduce a new change of coordinates, replacing the (potentially combinatorially growing) triangulation on the fly of \cite{vasilopoulos_pavlakos_schmeckpeper_daniilidis_koditschek_2019} with a fixed convex decomposition \cite{greene-1983} for each catalogued obstacle and revisit the prior hybrid dynamics convergence result \cite{vasilopoulos_pavlakos_schmeckpeper_daniilidis_koditschek_2019} to once again guarantee obstacle free geometric convergence. However, this streamlined computation, enabling full realtime integration of the Semantic SLAM engine, now allows us to react logically as well as geometrically within unexplored environments. In turn, realtime semantics combined with human recognition capability motivates the statement and proof of new rigorous guarantees for the robots to track suitably non-adversarial (see Definition \ref{definition:non_adversarial}) moving targets, while maintaining collision avoidance guarantees.

\subsubsection{Empirical Contributions}

We suggest the utility of the proposed architecture with a numerical study including comparisons with a state-of-the-art dynamic replanning algorithm \cite{otte-2015}, and physical implementation on both a wheeled and legged platform in highly varied environments (cluttered outdoor and indoor spaces including sunlight-flooded linoleum floors as well as featureless carpeted  hallways). Targets are robustly followed up to speeds amenable to the perceptual pipeline's tracking rate. Importantly, the semantic capabilities of the perceptual pipeline are exploited to introduce more complex task logic (e.g., track a given target unless encountering a specific human gesture).


\subsection{Organization of the Paper and Supplementary Material}
After stating the problem, summarizing our solution and introducing technical notation in Section \ref{sec:problemformulation}, Section \ref{sec:diffeomorphism} describes the diffeomorphism between the mapped and model spaces, and Section \ref{sec:reactiveplanning} includes our main formal results. Section \ref{sec:simulations} and Section \ref{sec:experiments} continue with our numerical and experimental studies, and Section \ref{sec:conclusion} concludes with ideas for future work. The supplementary video submission provides visual context for our empirical studies; we also include pointers to open-source software implementations, including both the MATLAB simulation package\footnote{\url{https://github.com/KodlabPenn/semnav_matlab}}, and the ROS-based controller\footnote{\url{https://github.com/KodlabPenn/semnav}}, in C++ and Python.

\section{PROBLEM FORMULATION AND APPROACH}
\label{sec:problemformulation}

\subsection{Problem Formulation}
\label{subsec:problem_formulation}

As in \cite{vasilopoulos_koditschek_WAFR2018,vasilopoulos_pavlakos_schmeckpeper_daniilidis_koditschek_2019}, we consider a robot with radius $\robotradius$, centered at $\robotposition \in \mathbb{R}^2$, navigating a compact, polygonal, potentially non-convex workspace $\workspace \subset \mathbb{R}^2$, with known boundary $\partial \workspace$, towards a target $\goalposition \in \workspace$. The robot is assumed to possess a sensor with fixed range $\sensorrange$, for recognizing ``familiar'' objects and estimating distance to nearby obstacles\footnote{For our hardware implementation, this idealized sensor is reduced to a combination of a LIDAR for distance measurements to obstacles and a monocular camera for object recognition and pose identification.}. We define the {\it enclosing workspace}, as the convex hull of the closure of the workspace $\workspace$, i.e., $\enclosingworkspace := \left\{ \robotposition \in \mathbb{R}^2 \, | \, \robotposition \in \text{Conv}(\overline{\workspace}) \right\}$.

The workspace is cluttered by a finite but unknown number of disjoint obstacles, denoted by $\obstacleset:=\{\obstacle_1,\obstacle_2,\ldots\}$, which might also include non-convex ``intrusions'' of the boundary of the physical workspace $\workspace$ into $\enclosingworkspace$. As in \cite{Arslan_Koditschek_2018,vasilopoulos_pavlakos_schmeckpeper_daniilidis_koditschek_2019}, we define the \textit{freespace} $\freespace$ as the set of collision-free placements for the closed ball $\ballclosure{\robotposition}{\robotradius}$ centered at $\robotposition$ with radius $\robotradius$, and the \textit{enclosing freespace}, $\enclosingfreespace$, as $\enclosingfreespace := \left\{ \robotposition \in \mathbb{R}^2 \, | \, \robotposition \in \text{Conv}(\overline{\freespace}) \right\}$.

Although none of the positions of any obstacles in $\obstacleset$ are \`{a}-priori known, a subset $\knownobstacleset:=\{\knownobstacle_i\}_{i \in \knownobstaclesetindex} \subseteq \obstacleset$ of these obstacles, indexed by $\knownobstaclesetindex:=\{1,\ldots,\knownobstaclecardinality\} \subset \mathbb{N}$, is assumed to be ``familiar'' in the sense of having a known, readily recognizable polygonal geometry, that the robot can instantly identify and localize. The remaining obstacles in $\unknownobstacleset:=\obstacleset\backslash\knownobstacleset$, indexed by $\unknownobstaclesetindex:=\{1,\ldots,\unknownobstaclecardinality\} \subset \mathbb{N}$, are assumed to be strongly convex according to \cite[Assumption 2]{Arslan_Koditschek_2018}, but are otherwise completely unknown to the robot.

To simplify the notation, we dilate each obstacle by $\robotradius$, and assume that the robot operates in the freespace $\mathcal{F}$. We denote the set of dilated obstacles derived from $\obstacleset, \knownobstacleset$ and $\unknownobstacleset$, by $\obstaclesetdilated, \knownobstaclesetdilated$ and $\unknownobstaclesetdilated$ respectively. Then, similarly to \cite{rimon1992,vasilopoulos_koditschek_WAFR2018,vasilopoulos_pavlakos_schmeckpeper_daniilidis_koditschek_2019}, we describe each polygonal obstacle $\knownobstacledilated_i \in \knownobstaclesetdilated \subseteq \obstaclesetdilated$ by an \textit{obstacle function}, $\implicitgeneric_i(\robotposition)$, a real-valued map providing an implicit representation of the form $\knownobstacledilated_i = \{ \robotposition \in \mathbb{R}^2 \, | \, \implicitgeneric_i(\robotposition) \leq 0 \}$ that the robot can construct online after it has localized $\knownobstacledilated_i$, following \cite{shapiro2007}. We also require the following separation assumptions.

\begin{assumption} \label{assumption:separations}
\begin{enumerate}
    \item Each obstacle $\unknownobstacledilated_i \in \unknownobstaclesetdilated$ has a positive clearance $d(\unknownobstacledilated_i,\unknownobstacledilated_j) > 0$ from any obstacle $\unknownobstacledilated_j \in \unknownobstaclesetdilated$, $j \neq i$. Also, $d(\unknownobstacledilated_i, \partial \freespace) > 0$, $\forall C_i \in \unknownobstaclesetdilated$.
    \item For each $\knownobstacledilated_i \in \knownobstaclesetdilated$, there exists $\betaclearance_i>0$ such that the set $S_{\implicitgeneric_i} := \{\robotposition \, | \, \implicitgeneric_i(\robotposition) \leq \betaclearance_i \}$ has a positive clearance $d(S_{\implicitgeneric_i},\unknownobstacledilated) > 0$ from any obstacle $\unknownobstacledilated \in \unknownobstaclesetdilated$.
\end{enumerate}
\end{assumption}

Based on these assumptions and considering first-order dynamics $\dot{\robotposition} = \controlfullyactuated(\robotposition)$, the problem consists of finding a Lipschitz continuous controller $\controlfullyactuated:\freespace \rightarrow \mathbb{R}^2$, that leaves the path-connected freespace $\freespace$ positively invariant and steers the robot to the (possibly moving) goal $\goalposition \in \freespace$.

\subsection{Overview of the Solution}

We solve the aforementioned problem by interpolating a sequence of spaces between the physical and a topologically equivalent but geometrically simple model space, designing our control input in the model space, and transforming this input through the inverse of the diffeomorphism between the mapped and the model space (Section \ref{sec:diffeomorphism}) to find the commands in the physical space (Section \ref{sec:reactiveplanning}).

To this end, we arrive at the central formal results (Proposition \ref{proposition:diffeo_purging}, Theorems \ref{theorem:control_fullyactuated}-\ref{theorem:control_se2}) by employing the smooth ``switch'' and ``deforming factor'' construction (see \eqref{eq:sigma_ji}, \eqref{eq:deforming_factor_purging}), integrated into the prior hybrid systems navigational framework from \cite{vasilopoulos_pavlakos_schmeckpeper_daniilidis_koditschek_2019}, that had, in turn, relied on the method for generating differential drive inputs from \cite{vasilopoulos_koditschek_WAFR2018}.

\subsection{Environment Representation and Technical Notation}
\label{subsec:environment}

Here, we establish notation for the four distinct representations of the environment that we will refer to as {\it planning spaces} \cite{vasilopoulos_koditschek_WAFR2018,vasilopoulos_pavlakos_schmeckpeper_daniilidis_koditschek_2019}, as shown in Fig. \ref{fig:diffeo_idea}. The robot navigates the physical space and discovers obstacles, that are dilated by the robot radius $\robotradius$ and stored in the semantic space. Potentially overlapping obstacles in the semantic space are subsequently consolidated in real time to form the mapped space. A change of coordinates from this space is then employed to construct a geometrically simplified (but topologically equivalent) model space, by merging familiar obstacles overlapping with the boundary of the enclosing freespace $\partial \enclosingfreespace$ to $\partial \enclosingfreespace$, deforming other familiar obstacles to disks, and leaving unknown obstacles intact.

\subsubsection{Physical Space}
\label{subsec:physical_space}
The \textit{physical space} is a description of the actual workspace $\enclosingworkspace$ punctured with the obstacles $\obstacleset$, and is inaccessible to the robot. The robot navigates this space toward $\goalposition$, and discovers and localizes new obstacles along the way. We denote by $\knownobstaclesetphysical := \{ \knownobstacle_i\}_{i \in \hybridmode} \subseteq \knownobstacleset$ the set of physically ``instantiated'' familiar objects, i.e., all objects whose geometry and pose are known to the robot either beforehand (when considering, e.g., a known wall layout), or by performing online localization at execution time, using semantic mapping. This set is indexed by a set $\hybridmode \subseteq \knownobstaclesetindex$, also defining the modes of our hybrid controller (Section \ref{sec:reactiveplanning}).

\subsubsection{Semantic Space}
\label{subsec:semantic_space}
The \textit{semantic space} $\freespacesemantic$ describes the robot's constantly updated information about the environment, from the observable portions of a subset of unrecognized obstacles, together with the $|\hybridmode|$ instantiated familiar obstacles. We denote the \textit{set of unrecognized obstacles in the semantic space} by $\unknownobstaclesetdilatedsemantic:= \{ \unknownobstacledilated_i \}_{i \in \unknownobstaclesetdilatedsemanticindex}$, indexed by $\unknownobstaclesetdilatedsemanticindex \subseteq \unknownobstaclesetindex$, and the \textit{set of familiar obstacles in the semantic space} by $\knownobstaclesetdilatedsemantic := \bigsqcup_{i \in \hybridmode} \knownobstacledilated_i$. Here the robot is treated as a single point.

\subsubsection{Mapped Space}
\label{subsec:mapped_space}
The semantic space does not explicitly contain any topological information about the explored environment, since Assumption \ref{assumption:separations} does not exclude overlaps between obstacles in $\knownobstaclesetdilated$.Hence, we form the \textit{mapped space}, $\freespacemapped$, by taking unions of elements of $\knownobstaclesetdilatedsemantic$, making up a new set of \textit{consolidated familiar obstacles} $\knownobstaclesetdilatedmapped := \{ \knownobstacledilated_i\}_{i \in \knownobstaclesetdilatedmappedindex}$ indexed by $\knownobstaclesetdilatedmappedindex$, with $|\knownobstaclesetdilatedmappedindex| \leq |\hybridmode|$, along with copies of the unknown obstacles (i.e., $\unknownobstaclesetdilatedmapped := \unknownobstaclesetdilatedsemantic$).

Next, for any connected component $\knownobstacledilated$ of $\knownobstaclesetdilatedmapped$ that intersects the boundary of the enclosing freespace $\partial \enclosingfreespace$, we take $\knownobstacledilatedmappedintrusion:= \knownobstacledilated \cap \enclosingfreespace$ and include $\knownobstacledilatedmappedintrusion$ in a new set $\knownobstaclesetdilatedmappedintrusion$, indexed by $\knownobstaclesetdilatedmappedintrusionindex$. The rest of the connected components in $\knownobstaclesetdilatedmapped$, which do not intersect $\partial \enclosingfreespace$, are included in a set $\knownobstaclesetdilatedmappeddisk$, indexed by $\knownobstaclesetdilatedmappeddiskindex$. The idea is that obstacles in $\knownobstaclesetdilatedmappedintrusion$ should be merged to $\partial \enclosingfreespace$, and obstacles in $\knownobstaclesetdilatedmappeddisk$ should be deformed to disks, since $\freespacemapped$ and $\freespacemodel$ need to be diffeomorphic.

\subsubsection{Model Space}
\label{subsec:model_space}
The \textit{model space} $\freespacemodel$ is a topologically equivalent but geometrically simplified version of the mapped space $\freespacemapped$. It has the same boundary as $\enclosingfreespace$ and consists of copies of the sensed fragments of the $|\unknownobstaclesetdilatedsemanticindex|$ unrecognized visible obstacles in $\unknownobstaclesetdilatedmapped$, and a collection of $|\knownobstaclesetdilatedmappeddiskindex|$ Euclidean disks corresponding to the $|\knownobstaclesetdilatedmappeddiskindex|$ consolidated obstacles in $\knownobstaclesetdilatedmappeddisk$ that are deformed to disks. Obstacles in $\knownobstaclesetdilatedmappedintrusion$ are merged into $\partial \enclosingfreespace$, to make $\freespacemapped$ and $\freespacemodel$ topologically equivalent, through a map $\diffeo$, described next.
\begin{figure*}[ht]
\captionsetup{width=\linewidth,font=footnotesize}
\centering
\includegraphics[width=1.0\textwidth]{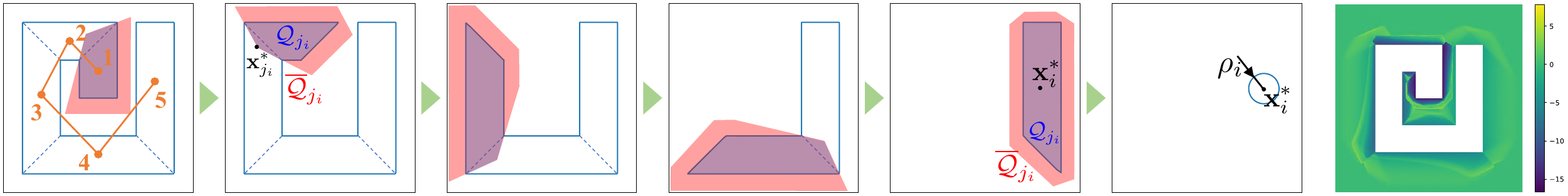}
\caption{Diffeomorphism construction via direct convex decomposition: Any arbitrary convex decomposition (e.g., \cite{greene-1983}) defines a tree $\polygontree{\knownobstacledilated_i}:=(\polygonvertices{\knownobstacledilated_i},\polygonedges{\knownobstacledilated_i})$ (left), which induces the sequence of purging transformations that map the polygon's boundary and exterior to the boundary and exterior of an equivalent disk. The purging transformation for each convex piece $j_i \in \polygonvertices{\knownobstacledilated_i}$ is defined by a pair of convex polygons $\innerpolygon{j_i}, \outerpolygon{j_i}$ that limit the effect of the diffeomorphism to a neighborhood of $j_i$. The final map is guaranteed to be smooth, as shown by a visualization of its determinant in logarithmic scale (right).}
\label{fig:purging_transformation}
\vspace{-14pt}
\end{figure*}

\section{DIFFEOMORPHISM CONSTRUCTION}
\label{sec:diffeomorphism}

Here, we describe our method of constructing the diffeomorphism, $\diffeo$, between $\freespacemapped$ and $\freespacemodel$. We assume that the robot has recognized and localized the $|\knownobstaclesetdilatedmappedindex|$ obstacles in $\knownobstaclesetdilatedmapped$, and has, therefore, identified obstacles to be merged to the boundary of the enclosing freespace $\partial \enclosingfreespace$, stored in $\knownobstaclesetdilatedmappedintrusion$, and obstacles to be deformed to disks, stored in $\knownobstaclesetdilatedmappeddisk$.

\subsection{Obstacle Representation and Convex Decomposition}
As a natural extension to doubly reactive algorithms for environments cluttered with convex obstacles \cite{Arslan_Koditschek_2018,Paternain_Koditschek_Ribeiro_2017}, we assume that the robot has access to the convex decomposition of each obstacle $\knownobstacledilated \in \knownobstaclesetdilatedmapped$. For polygons without holes, we are interested in decompositions that do not introduce Steiner points (i.e., additional points except for the polygon vertices), as this guarantees the dual graph of the convex partition to be a tree. Here, we acquire this convex decomposition using Greene's method \cite{greene-1983} and its C++ implementation in CGAL \cite{cgal}, operating in $\mathcal{O}(r^2n^2)$ time, with $n$ the number of polygon vertices $r$ the number of reflex vertices. Other algorithms \cite{lien-amato-2004} could be used as well, such as Keil's decomposition algorithm \cite{keil-snoeyink-2002,keil-convex-decomposition}, operating in $\mathcal{O}(r^2n^2\log{n})$ time.

As shown in Fig. \ref{fig:purging_transformation}, convex partioning results in a {\it tree of convex polygons} $\polygontree{\knownobstacledilated_i}:=(\polygonvertices{\knownobstacledilated_i},\polygonedges{\knownobstacledilated_i})$ corresponding to $\knownobstacledilated_i$, with $\polygonvertices{\knownobstacledilated_i}$ a set of vertices identified with convex polygons (i.e., vertices of the dual of the formal partition) and $\polygonedges{\knownobstacledilated_i}$ a set of edges encoding polygon adjacency. Therefore, we can pick any polygon as root and construct $\polygontree{\knownobstacledilated_i}$ based on the adjacency properties induced by the dual graph of the decomposition, as shown in Fig. \ref{fig:purging_transformation}. If $\knownobstacledilated_i \in \knownobstaclesetdilatedmappeddisk$, we pick as root the polygon with the largest surface area, whereas if $\knownobstacledilated_i \in \knownobstaclesetdilatedmappedintrusion$, we pick as root any polygon adjacent to $\partial \enclosingfreespace$.


\subsection{The Map Between the Mapped and the Model Space}
As shown in Fig. \ref{fig:purging_transformation}, the map $\diffeo$ between the mapped and the model space is constructed in several steps, involving the successive application of purging transformations by composition, during execution time, for all leaf polygons of all obstacles $\knownobstacledilated$ in $\knownobstaclesetdilatedmappedintrusion$ and $\knownobstaclesetdilatedmappeddisk$, in any order, until their root polygons are reached. We denote by $\freespacemappedhat$ this final intermediate space, where all obstacles in $\freespacemapped$ have been deformed to their root polygons. We denote by $\freespacemappedpurging{j_i}$ and $\freespacemappedpurging{p(j_i)}$ the intermediate spaces before and after the purging transformation of leaf polygon $j_i \in \polygonvertices{\knownobstacledilated_i}$ respectively.

We begin our exposition with a description of the purging transformation $\diffeopurging{j_i}: \freespacemappedpurging{j_i} \rightarrow \freespacemappedpurging{p(j_i)}$ that maps the boundary of a leaf polygon $j_i \in \polygonvertices{\knownobstacledilated_i}$ onto the boundary of its parent, $p(j_i)$, and continue with a description of the map $\diffeoroot:\freespacemappedhat \rightarrow \freespacemodel$ that maps the boundaries of root polygons of obstacles in $\knownobstaclesetdilatedmappedintrusion$ and $\knownobstaclesetdilatedmappeddisk$ to $\enclosingfreespace$ and the corresponding disks in $\freespacemodel$ respectively.

\subsubsection{The map between $\freespacemappedpurging{j_i}$ and $\freespacemappedpurging{p(j_i)}$}
\label{subsubsec:purging}
We first find admissible centers $\diffeocenter{j_i}$, and polygonal collars $\outerpolygon{j_i}$, that encompass the actual polygon $\innerpolygon{j_i}$, and limit the effect of the purging transformation in their interior, while keeping its value equal to the identity everywhere else (see Fig. \ref{fig:purging_transformation}).

\begin{definition}
\label{definition:center}
An admissible center for the purging transformation of the leaf polygon $j_i \in \polygonvertices{\knownobstacledilated_i}$, denoted by $\diffeocenter{j_i}$, is a point in $p(j_i)$ such that the polygon $\innerpolygon{j_i}$ with vertices the original vertices of $j_i$ and $\diffeocenter{j_i}$ is convex.
\end{definition}

\begin{definition}
\label{definition:collars}
An admissible polygonal collar for the purging transformation of the leaf polygon $j_i$ is a convex polygon $\outerpolygon{j_i}$ such that:
\begin{enumerate}
    \item $\outerpolygon{j_i}$ does not intersect the interior of any polygon $k \in \polygonvertices{\knownobstacledilated}$ with $k \neq j_i, p(j_i), \forall \knownobstacledilated \in \freespacemappedpurging{j_i}$, or any $\unknownobstacledilated \in \unknownobstaclesetdilatedmapped$.
    \item $\innerpolygon{j_i} \subset \outerpolygon{j_i}$, and $\outerpolygon{j_i} \backslash \innerpolygon{j_i} \subset \freespacemappedpurging{j_i}$.
\end{enumerate}
\end{definition}

Examples are shown in Fig. \ref{fig:purging_transformation}. As in \cite{vasilopoulos_pavlakos_schmeckpeper_daniilidis_koditschek_2019}, we also construct implicit functions $\innerpolygonimplicit{j_i}(\robotposition), \outerpolygonimplicit{j_i}(\robotposition)$ corresponding to the leaf polygon $j_i \in \polygonvertices{\knownobstacledilated_i}$ such that $\innerpolygon{j_i} = \{\robotposition \in \mathbb{R}^2 \, | \, \innerpolygonimplicit{j_i}(\robotposition) \leq 0\}$ and $\outerpolygon{j_i} = \{\robotposition \in \mathbb{R}^2 \, | \, \outerpolygonimplicit{j_i}(\robotposition) \geq 0\}$, using tools from \cite{shapiro2007}. 

Based on these definitions, we construct the {\it $C^\infty$ switch of the purging transformation for the leaf polygon $j_i \in \polygonvertices{\knownobstacledilated_i}$} as a function $\switch{j_i}:\freespacemappedpurging{j_i} \rightarrow \mathbb{R}$, equal to 1 on the boundary of $\innerpolygon{j_i}$, equal to 0 outside $\outerpolygon{j_i}$ and smoothly varying (except the polygon vertices) between 0 and 1 everywhere else (see \eqref{eq:sigma_ji} in Appendix \ref{appendix:proofs}). Finally, we define the {\it deforming factors} as the functions $\deformingfactor{j_i}:\freespacemappedpurging{j_i} \rightarrow \mathbb{R}$, responsible for mapping the boundary of the leaf polygon $j_i$ onto the boundary of its parent $p(j_i)$ (see \eqref{eq:deforming_factor_purging} in Appendix \ref{appendix:proofs}). We can now construct the map between $\freespacemappedpurging{j_i}$ and $\freespacemappedpurging{p(j_i)}$ as in \cite{vasilopoulos_koditschek_WAFR2018,vasilopoulos_pavlakos_schmeckpeper_daniilidis_koditschek_2019}
\begin{equation}
    \diffeopurging{j_i}(\robotposition) := \switch{j_i}(\robotposition) \left( \diffeocenter{j_i} + \deformingfactor{j_i}(\robotposition)(\robotposition-\diffeocenter{j_i}) \right) + \left(1-\switch{j_i}(\robotposition) \right) \robotposition \nonumber \label{eq:map_purging}
\end{equation}

\begin{proposition}
\label{proposition:diffeo_purging}
The map $\diffeopurging{j_i}$ is a $C^\infty$ diffeomorphism between $\freespacemappedpurging{j_i}$ and $\freespacemappedpurging{p(j_i)}$ away from the polygon vertices of $j_i$, none of which lies in the interior of $\freespacemappedpurging{j_i}$.
\end{proposition}
\begin{proof}
Included in Appendix \ref{appendix:proofs}.
\end{proof}

We denote by $\diffeocomposition:\freespacemapped \rightarrow \freespacemappedhat$ the map between $\freespacemapped$ and $\freespacemappedhat$, arising from the composition of purging transformations $\diffeopurging{j_i}:\freespacemappedpurging{j_i} \rightarrow \freespacemappedpurging{p(j_i)}$.

\subsubsection{The Map Between $\freespacemappedhat$ and $\freespacemodel$}

Here, for each root polygon $r_i$, we define the polygonal collar and the {\it $C^\infty$ switch of the transformation} $\switch{r_i}:\freespacemappedhat \rightarrow \freespacemapped$ as in Definition \ref{definition:collars} and \eqref{eq:sigma_ji} (see Appendix \ref{appendix:proofs}) respectively, and we distinguish between obstacles in $\knownobstaclesetdilatedmappedintrusion$ and in $\knownobstaclesetdilatedmappeddisk$ for the definition of the centers as follows (see Fig. \ref{fig:purging_transformation}).
\begin{definition}
    An admissible center for the transformation of:
    \begin{enumerate}
        \item the root polygon $r_i$, corresponding to $\knownobstacledilated_i \in \knownobstaclesetdilatedmappeddisk$, is a point $\diffeocenter{i}$ in the interior of $r_i$ (here identified with $\innerpolygon{r_i}$).
        \item the root polygon $r_i$, corresponding to $\knownobstacledilated_i \in \knownobstaclesetdilatedmappedintrusion$, is a point $\diffeocenter{i} \in \mathbb{R}^2 \backslash \enclosingfreespace$, such that the polygon $\innerpolygon{r_i}$ with vertices the original vertices of $r_i$ and $\diffeocenter{i}$ is convex.
    \end{enumerate}
    \label{definition:center_root}
\end{definition}

Finally, we define the {\it deforming factors} $\deformingfactor{r_i}:\freespacemappedhat \rightarrow \mathbb{R}$ as in Section \ref{subsubsec:purging} for obstacles in $\knownobstaclesetdilatedmappedintrusion$, and as the function $\deformingfactor{r_i}(\robotposition) := \frac{\diffeoradius{i}}{||\robotposition-\diffeocenter{i}||}$ for obstacles in $\knownobstaclesetdilatedmappeddisk$ (see Fig. \ref{fig:purging_transformation}). We construct the map between $\freespacemappedhat$ and $\freespacemodel$ as
\begin{equation}
    \diffeoroot(\robotposition):=\sum_{i \in \knownobstaclesetdilatedmappedintrusionindex \cup \knownobstaclesetdilatedmappeddiskindex} \switch{r_i}(\robotposition) \left[ \diffeocenter{i} + \deformingfactor{r_i}(\robotposition)(\robotposition-\diffeocenter{i}) \right] + \switchresidual(\robotposition) \robotposition \nonumber \label{eq:map_root}
\end{equation}
with $\switchresidual(\robotposition) := 1-\sum_{i \in \knownobstaclesetdilatedmappedintrusionindex \cup \knownobstaclesetdilatedmappeddiskindex} \switch{r_i}(\robotposition)$. It should be noted that Definitions \ref{definition:collars} and \ref{definition:center_root} guarantee that, at any point in the workspace, at most one switch $\switch{r_i}$ will be greater than zero which, in turn, guarantees that the diffeomorphism computation is essentially ``local'', and allows scaling to multiple obstacles in the mapped space $\freespacemapped$.

We can similarly arrive at the following result.
\begin{proposition}
\label{proposition:diffeo_root}
The map $\diffeoroot$ is a $C^\infty$ diffeomorphism between $\freespacemappedhat$ and $\freespacemodel$ away from any sharp corners, none of which lie in the interior of $\freespacemappedhat$.
\end{proposition}

\subsubsection{The Map Between $\freespacemapped$ and $\freespacemodel$}
Based on the construction of $\diffeocomposition:\freespacemapped \rightarrow \freespacemappedhat$ and $\diffeoroot:\freespacemappedhat \rightarrow \freespacemodel$, we can finally write the map between the mapped space and the model space as the function $\diffeo : \freespacemapped \rightarrow \freespacemodel$ given by $\diffeo(\robotposition) = \diffeoroot \circ \diffeocomposition(\robotposition)$. Since both $\diffeocomposition$ and $\diffeoroot$ are $C^\infty$ diffeomorphisms away from sharp corners, it is straightforward to show that the map $\diffeo$ is a $C^\infty$ diffeomorphism between $\freespacemapped$ and $\freespacemodel$ away from any sharp corners, none of which lie in the interior of $\freespacemapped$.
\begin{figure*}[t]
\captionsetup{width=\linewidth,font=footnotesize}
\centering
\includegraphics[width=1.0\textwidth]{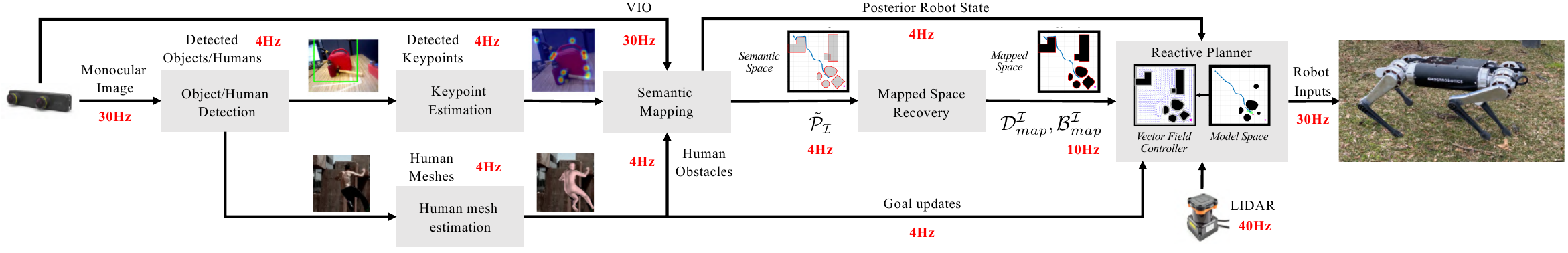}
\caption{The online reactive planning architecture: Advancing beyond \cite{vasilopoulos_pavlakos_schmeckpeper_daniilidis_koditschek_2019}, camera output is run through a perceptual pipeline incorporating three separate neural networks (run onboard at 4Hz) whose function is to: (a) detect familiar obstacles and humans \cite{yolov3}; (b) localize corresponding semantic keypoints \cite{Pavlakos2017}; and (c) perform a 3D human mesh estimation \cite{kolotouros2019learning}. Keypoint locations on the image, other detected geometric features, and an egomotion estimate provided by visual inertial odometry are used by the semantic mapping module \cite{Bowman2017} to give updated robot ($\robotposition$) and obstacle poses ($\knownobstaclesetphysical$). The reactive planner, now streamlined to run onboard at 3x the rate of the corresponding module in \cite{vasilopoulos_pavlakos_schmeckpeper_daniilidis_koditschek_2019}, merges consolidated obstacles in $\knownobstaclesetdilatedmappeddisk, \knownobstaclesetdilatedmappedintrusion$ (recovered from $\knownobstaclesetphysical$), along with LIDAR data for unknown obstacles, to provide the robot inputs and close the control loop. In this new architecture, the estimated human meshes are used to update the target's position in the reported human tracking experiments, detect a specific human gesture or pose related to the experiment's semantics, or (optionally) introduce additional obstacles in the semantic mapping module for some out-of-scope experiments.}
\label{fig:algorithm}
\vspace{-14pt}
\end{figure*}

\section{REACTIVE PLANNING ALGORITHM}
\label{sec:reactiveplanning}

The analysis in Section \ref{sec:diffeomorphism} describes the diffeomorphism construction between $\freespacemapped$ and $\freespacemodel$ for a given index set $\hybridmode$ of instantiated familiar obstacles. However, the onboard sensor might incorporate new obstacles in the semantic map, updating $\hybridmode$. Therefore, as in \cite{vasilopoulos_pavlakos_schmeckpeper_daniilidis_koditschek_2019}, we give a hybrid systems description of our reactive controller, where each mode is defined by an index set $\hybridmode \in 2^{\knownobstaclesetindex}$ of familiar obstacles stored in the semantic map, the guards describe the sensor trigger events where a previously ``unexplored'' obstacle is discovered and incorporated in the semantic map (thereby changing $\knownobstaclesetdilatedmapped$, and $\knownobstaclesetdilatedmappeddisk$, $\knownobstaclesetdilatedmappedintrusion$) \cite[Eqns. (31),(35)]{vasilopoulos_pavlakos_schmeckpeper_daniilidis_koditschek_2019}, and the resets describe transitions to new modes that are equal to the identity in the physical space, but might result in discrete ``jumps'' of the robot position in the model space \cite[Eqns. (32), (36)]{vasilopoulos_pavlakos_schmeckpeper_daniilidis_koditschek_2019}. In this work, this hybrid systems structure is not modified, and we just focus on each mode $\hybridmode$ separately.

For a fully actuated particle with dynamics $\dot{\robotposition} = \controlfullyactuated(\robotposition), \controlfullyactuated \in \mathbb{R}^2$, the control law in each mode $\hybridmode$ is given as
\begin{equation}
    \controlfullyactuated^\hybridmode(\robotposition) = k \left[ D_\robotposition \diffeo \right]^{-1} \cdot \left(\controlfullyactuatedmodel^\hybridmode \circ \diffeo(\robotposition) \right) \label{eq:control_fullyactuated}
\end{equation}
with $D_\robotposition$ denoting the derivative operator with respect to $\robotposition$, and the control input in the model space given as \cite{Arslan_Koditschek_2018}
\begin{equation}
    \controlfullyactuatedmodel^\hybridmode(\robotpositionmodel) = - \left(\robotpositionmodel - \projection{\localfreespace{\robotpositionmodel}}{\goalpositionmodel} \right) \label{eq:control_fullyactuated_model}
\end{equation}
Here, $\robotpositionmodel = \diffeo(\robotposition) \in \hybridfreespacemodemodel$ and $\goalpositionmodel = \diffeo(\goalposition)$ denote the robot and goal position in the model space respectively, and $\projection{\localfreespace{\robotpositionmodel}}{\goalpositionmodel}$ denotes the projection onto the convex {\it local freespace} for $\robotpositionmodel$, $\localfreespace{\robotpositionmodel}$, defined as the Voronoi cell in \cite[Eqn. (25)]{Arslan_Koditschek_2018}, separating $\robotpositionmodel$ from all the model space obstacles (see Fig. \ref{fig:diffeo_idea}). We use the following definition to define a slowly moving, non-adversarial moving target.

\begin{definition}
    \label{definition:non_adversarial}
    The smooth function $\goalposition:\mathbb{R} \rightarrow \freespacemapped$ is a {\it non-adversarial target} if its model space velocity, given as $\goalpositionmodeldot := D_\robotposition \diffeo(\goalposition) \cdot \goalpositiondot$, always satisfies either $(\diffeo(\robotposition)-\diffeo(\goalposition))^\top \goalpositionmodeldot \geq 0$, or $||\goalpositionmodeldot|| \leq k \, \dfrac{||\diffeo(\robotposition) - \projection{\ball{\diffeo(\robotposition)}{0.5d(\diffeo(\robotposition),\partial \freespacemodel)}}{\diffeo(\goalposition)} ||^2}{||\diffeo(\robotposition)-\diffeo(\goalposition)||}$.
\end{definition}

Intuitively, this Definition requires the moving target to slow down when the robot gets too close to obstacles (i.e., when $d(\diffeo(\robotposition),\partial \freespacemodel)$ becomes small) or the target itself (i.e., when $\projection{\ball{\diffeo(\robotposition)}{0.5d(\diffeo(\robotposition),\partial \freespacemodel)}}{\diffeo(\goalposition)} = \diffeo(\goalposition)$), proportionally to the control gain $k$, unless the target approaches the robot (i.e., $(\diffeo(\robotposition)-\diffeo(\goalposition))^\top \goalpositionmodeldot \geq 0$). We use Definition \ref{definition:non_adversarial} to arrive at the following central result.

\begin{theorem}
    \label{theorem:control_fullyactuated}
    With $\hybridmode$ the terminal mode of the hybrid controller\footnote{The {\it terminal mode} of the hybrid system is indexed by the improper subset, $\hybridmode = \knownobstaclesetindex$, where all familiar obstacles in the workspace have been instantiated in the set $\knownobstaclesetdilatedsemantic$.\label{footnote:terminal_mode}}, the reactive controller in \eqref{eq:control_fullyactuated} leaves the freespace $\freespacemapped$ positively invariant, and:
    \begin{enumerate}
        \item tracks $\goalposition$ by not increasing $||\diffeo(\robotposition) - \diffeo(\goalposition)||$, if $\goalposition$ is a non-adversarial target (see Definition \ref{definition:non_adversarial}).
        \item asymptotically reaches a constant $\goalposition$ with its unique continuously differentiable flow, from almost any placement $\robotposition \in \freespacemapped$, while strictly decreasing $||\diffeo(\robotposition) - \diffeo(\goalposition)||$ along the way.
    \end{enumerate}
\end{theorem}
\begin{proof}
Included in Appendix \ref{appendix:proofs}.
\end{proof}

In \cite{vasilopoulos_koditschek_WAFR2018}, we extended our algorithm to differential drive robots, by constructing a smooth diffeomorphism $\diffeounicycle:\freespacemapped \times S^1 \rightarrow \hybridfreespacemodemodel \times S^1$ away from sharp corners. We summarize the details of the construction in Appendix \ref{appendix:differential_drive}, and present our main result below, whose proof follows similar patterns to that of Theorem \ref{theorem:control_fullyactuated} and is omitted for brevity.

\begin{theorem} \label{theorem:control_se2}
With $\hybridmode$ the terminal mode of the hybrid controller$^{\ref{footnote:terminal_mode}}$, the reactive controller for differential drive robots (see \eqref{eq:control_unicycle} in Appendix \ref{appendix:differential_drive}) leaves the freespace $\freespacemapped \times S^1$ positively invariant, and:
\begin{enumerate}
    \item tracks $\goalposition$ by not increasing $||\diffeo(\robotposition) - \diffeo(\goalposition)||$, if $\goalposition$ is a non-adversarial target (see Definition \ref{definition:non_adversarial}).
    \item asymptotically reaches a constant $\goalposition$ with its unique continuously differentiable flow, from almost any robot configuration in $\freespacemapped \times S^1$, without increasing $|| \diffeo(\robotposition)-\diffeo(\goalposition)||$ along the way.
\end{enumerate}
\end{theorem}
\section{NUMERICAL STUDIES}
\label{sec:simulations}

In this Section, we present numerical studies run in MATLAB using $\texttt{ode45}$, that illustrate our formal results. Our reactive controller is implemented in Python and communicates with MATLAB using the standard MATLAB-Python interface. For our numerical results, we assume perfect robot state estimation and localization of obstacles, using a fixed range sensor that can instantly identify and localize either the entirety of familiar obstacles that intersect its footprint, or the fragments of unknown obstacles within its range.

\subsection{Illustrations of the Navigation Framework}

\begin{figure}[t]
\captionsetup{width=\linewidth,font=footnotesize}
\centering
\includegraphics[width=0.7\columnwidth]{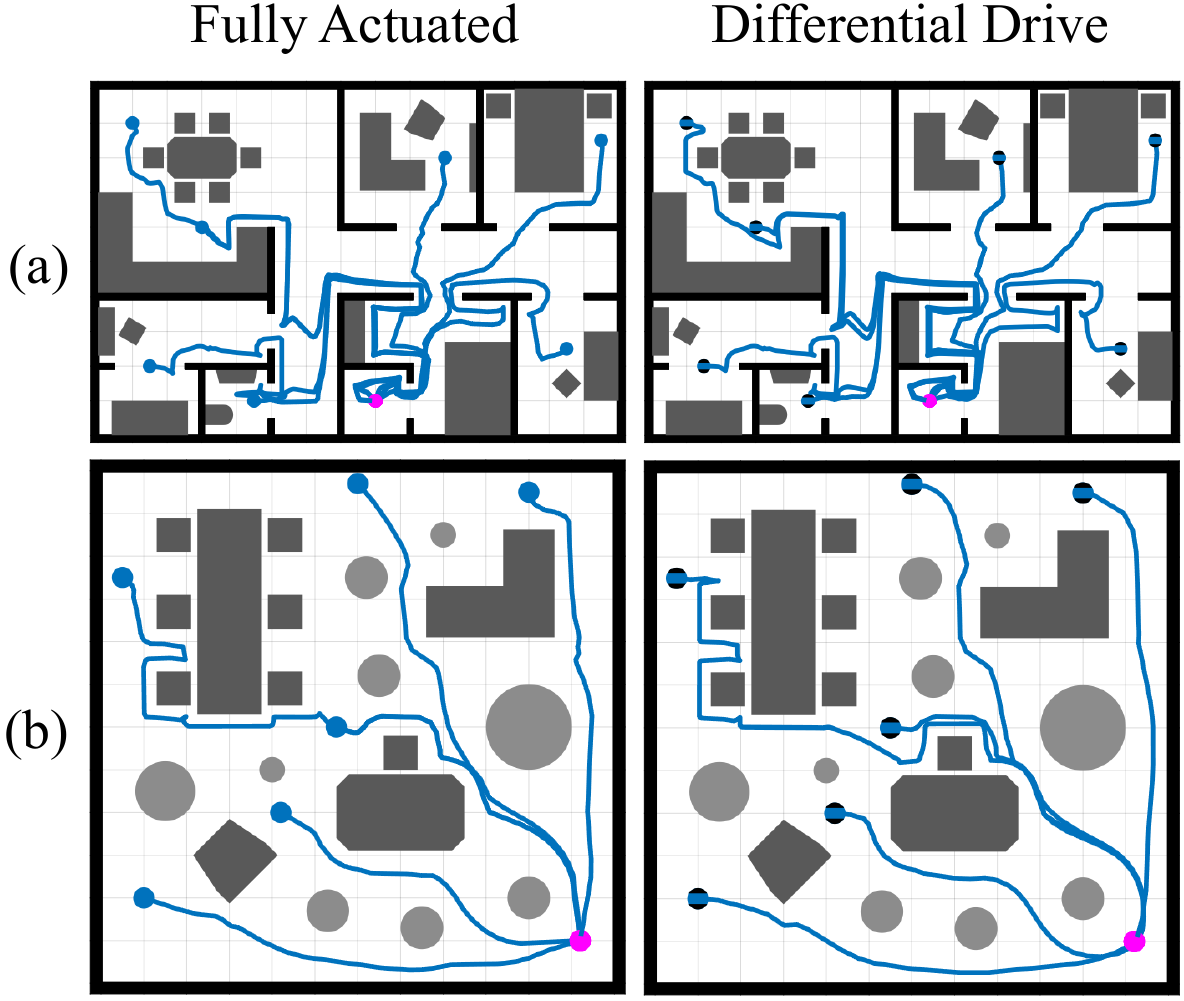}
\caption{Top: Navigation in an indoor layout cluttered with multiple familiar obstacles and previously unknown pose. - Bottom: Navigation in a room cluttered with known non-convex (dark grey) and unknown convex (light grey) obstacles. Simulations are run from different initial conditions.}
\label{fig:simulation_known_unknown}
\vspace{-18pt}
\end{figure}

We begin by illustrating the performance of our reactive planning framework in two different settings (Fig. \ref{fig:simulation_known_unknown}), for both a fully actuated and a differential drive robot, also included in the accompanying video submission. In the first case (Fig. \ref{fig:simulation_known_unknown}-a), the robot is tasked with moving to a predefined location in an environment resembling an apartment layout with known walls, cluttered with several familiar obstacles of unknown location and pose, from different initial conditions. In the second case (Fig. \ref{fig:simulation_known_unknown}-b), the robot navigates a room cluttered with both familiar and unknown obstacles from several initial conditions. In both cases, the robot avoids all the obstacles and safely converges to the target. The robot radius used in our simulation studies is 0.2m.


\subsection{Comparison with RRT$^X$ \cite{otte-2015}}

In the second set of numerical results, we compare our reactive controller with a state-of-the-art path replanning algorithm, RRT$^X$ \cite{otte-2015}. We choose to compare against this specific algorithm instead of another sampling-based method for static environments (e.g., RRT* \cite{karaman_frazzoli_ICRA2012}), since both our reactive controller and RRT$^X$ are dynamic in nature; they are capable of incorporating new information about the environment and modifying the robot's behavior appropriately. For our simulations, we assume that RRT$^X$ possesses the same sensory apparatus with our algorithm; an ``oracle'' that can instantly identify and localize nearby obstacles. The computed paths are then reactively tracked using \cite{arslan_kod_ICRA2017}.

\begin{figure}[t]
\captionsetup{width=\linewidth,font=footnotesize}
\centering
\includegraphics[width=0.9\columnwidth]{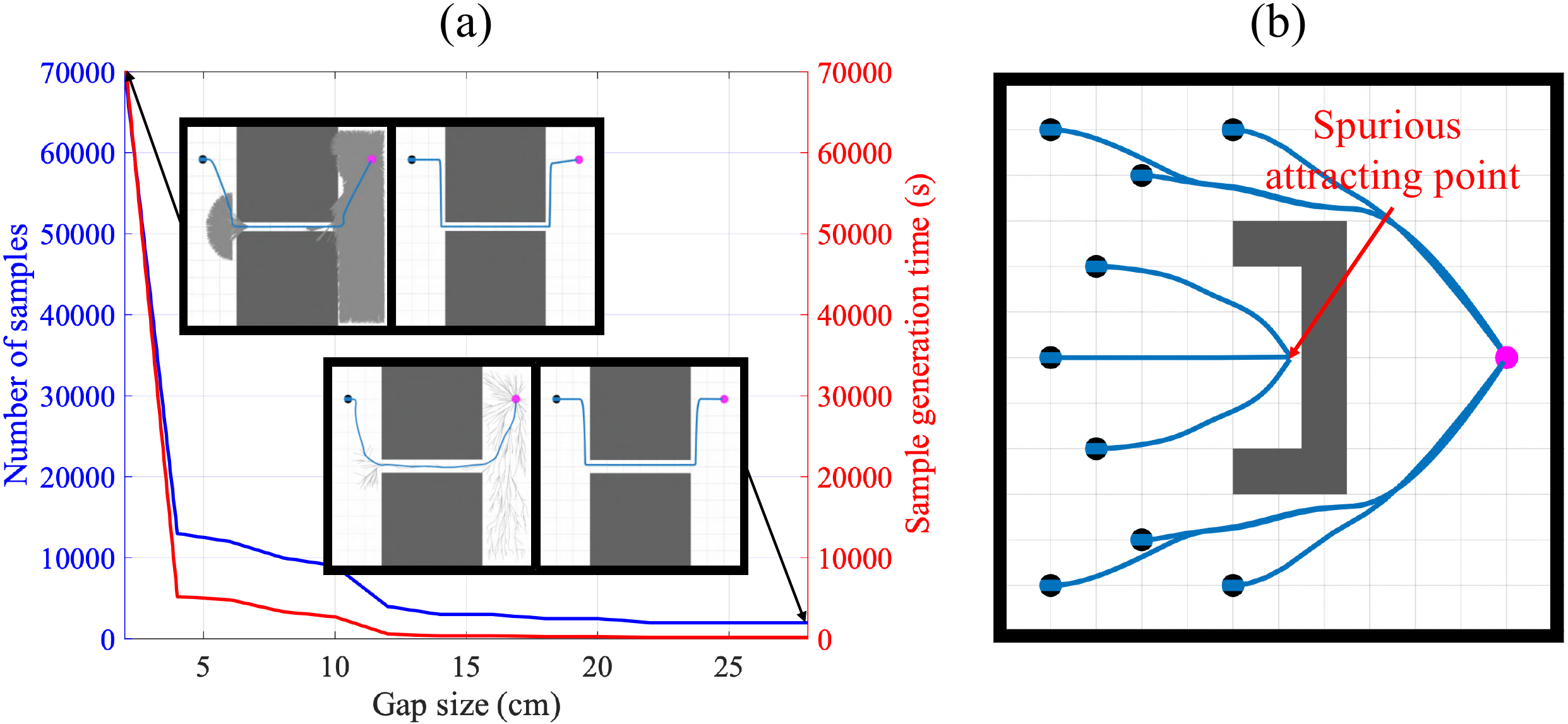}
\caption{(a) Minimum number of (offline computed) samples needed for successful online implementation of RRT$^X$ \cite{otte-2015} in an unexplored environment with two familiar obstacles forming a narrow passage. The number becomes increasingly large as the gap becomes smaller. The robot diameter is 50cm. (b) Illustration of a graceful failure of our proposed algorithm. The sole non-convex but unknown encountered obstacle creates a spurious attracting equilibrium state that traps a subset of initial conditions. However, collision avoidance is always guaranteed by the onboard sensor.}
\label{fig:rrt_narrow_passage}
\vspace{-18pt}
\end{figure}

Fig. \ref{fig:rrt_narrow_passage}-a exemplifies the (well-known \cite{hsu-2000}) performance degradation of RRT$^X$ in the presence of narrow passages: as the corridor narrows (while always larger than the robot's diameter), the minimum number of (offline-computed) samples needed for successful replanning and safe navigation increases in a nonlinear manner. In consequence of this dramatically growing time-to-completeness, our video demonstrates a potentially catastrophic failure of the associated replanner: in the presence of multiple narrow passages, it cycles repeatedly as it searches for possible alternative openings, before eventually (and only after increasingly protracted cycling) reporting failure (incorrectly) and halting. On the contrary, our algorithm always guarantees safe passage to the target through compliant environments -- and Fig. \ref{fig:rrt_narrow_passage}-b illustrates its graceful failure for settings that violate Assumption \ref{assumption:separations}. The non-compliant (novel but not convex) obstacle creates an attracting equilibrium state that traps a set of initial conditions whose area becomes arbitrarily large as its ``shadow'' (the associated basin of attraction) grows. However, the presence of a Lyapunov function precludes the possibility of any cycling behavior: failure to achieve the goal (and the diagnosis of a non-compliant environment) is readily identified.

\section{EXPERIMENTS}
\label{sec:experiments}

\begin{figure*}[!ht]
\captionsetup{width=\linewidth,font=footnotesize}
\centering
\includegraphics[width=1.0\textwidth]{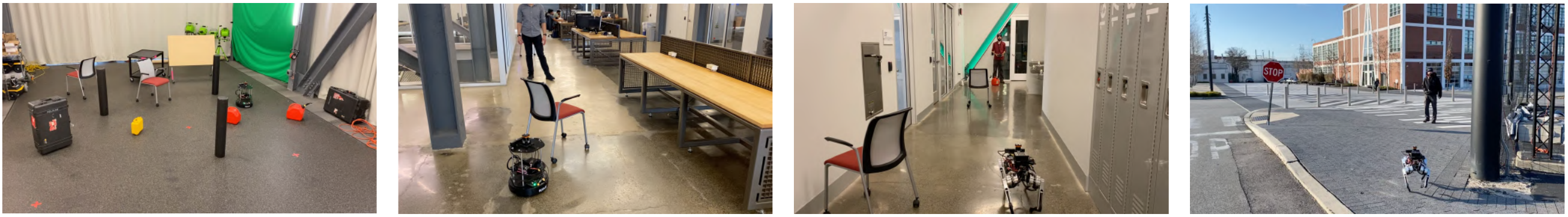}
\caption{Types of environments used in our experiments. Visual context is included in the supplementary video submission.}
\label{fig:environments}
\vspace{-14pt}
\end{figure*}

\begin{figure}[!ht]
\captionsetup{width=\linewidth,font=footnotesize}
\centering
\includegraphics[width=1.0\columnwidth]{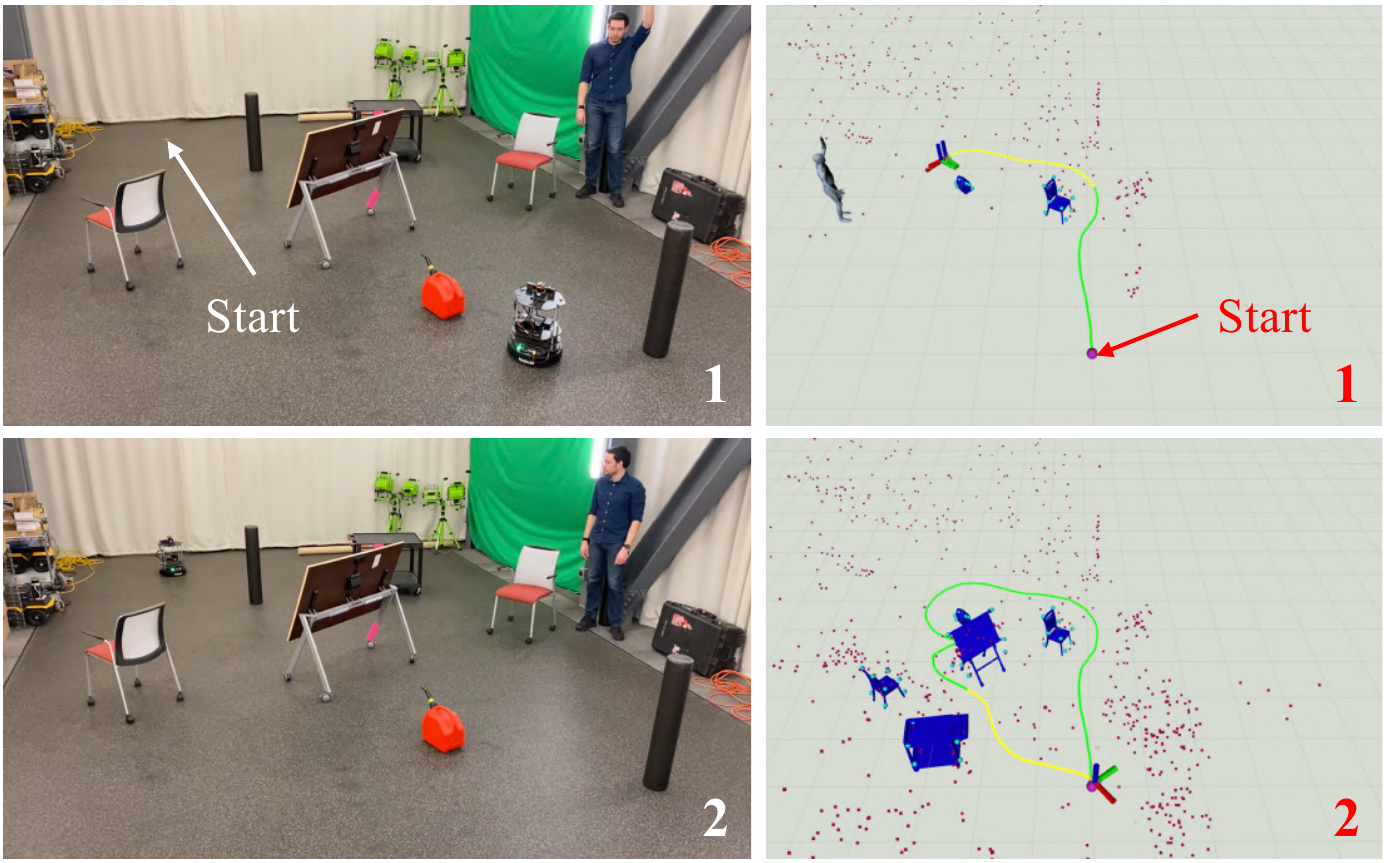}
\caption{Top: Turtlebot reactively follows a human until a stop gesture is given and detected -- Bottom: Turtlebot safely returns to its starting position.}
\label{fig:experiment_human_stop}
\vspace{-18pt}
\end{figure}

\subsection{Experimental Setup}

Our experimental layout is summarized in Fig. \ref{fig:algorithm}. Since the algorithms introduced in this paper take the form of first-order vector fields, we mainly use a quasi-static platform, the Turtlebot robot \cite{turtlebot} for our physical experiments. We suggest the robustness of these feedback controllers by performing several experiments on the more dynamic Ghost Spirit legged robot \cite{ghostspirit}, using a rough approximation to the quasi-static differential drive motion model. In both cases, the main computer is an Nvidia Jetson AGX Xavier GPU unit, responsible for running our perception and navigation algorithms, during execution time. This GPU unit communicates with a Hokuyo LIDAR, used to detect unknown obstacles, and a ZED Mini stereo camera, used for visual-inertial state estimation and for detecting humans and familiar obstacles. 

Our perception pipeline, run onboard the Nvidia Jetson AGX Xavier at 4Hz, supports the detection and 3D pose estimation of objects and humans, who, for the purposes of this paper, are used as moving targets. We use the YOLOv3 detector~\cite{yolov3} to detect 2D bounding boxes on the image which are then processed based on the class of the detected object. If one of the specified object classes is detected, then we follow the semantic keypoints approach of~\cite{Pavlakos2017} to estimate keypoints of the object on the image plane\footnote{Note that while both the YOLOv3 detector \cite{yolov3} and the keypoint estimation algorithm \cite{Pavlakos2017} are empirically very robust (e.g., particularly against partial occlusions), they could be easily replaced with other state-of-the-art algorithms that provide reasonable robustness against partial occlusions.}. The familiar object classes (as defined in Section \ref{sec:problemformulation}) used in our experiments are chair, table, ladder, cart, gascan and pelican case, although this dictionary can increase depending on the user's needs. The training data for the particular instances of interest are collected with a semi-automatic procedure, similarly to \cite{Pavlakos2017}. Given the bounding box and keypoint annotations for each image, the two networks are trained with their default configurations until convergence. On the other hand, if the bounding box corresponds to a person detection, then we use the approach of \cite{kolotouros2019learning}, that provides us with the 3D mesh of the person.

Our semantic mapping infrastructure relies on the algorithm presented in \cite{Bowman2017}, and is implemented in C++. This algorithm fuses inertial information (here provided by the position tracking implementation from StereoLabs on the ZED Mini stereo camera), geometric (i.e., geometric features on the 2D image), and semantic information (i.e., the detected keypoints and the associated object labels as described above) to give a posterior estimate for both the robot state and the associated poses of all tracked objects, by simultaneously solving the data association problem arising when several objects of the same class exist in the map.

Finally, our reactive controller, running online and onboard the Nvidia Jetson AGX Xavier GPU unit at 30Hz, is also implemented in C++ using Boost Geometry \cite{boost} for the underlying polygon operations, and communicates with our perception pipelines using ROS, as shown in Fig. \ref{fig:algorithm}.

\subsection{Empirical Results}

As also reported in the supplementary video, we distinguish between two classes of physical experiments in several different environments shown in Fig. \ref{fig:environments}; tracking either a predefined static target or a moving human, and tracking a given semantic target (e.g., approach a desired object).

\subsubsection{Geometric tracking of a (moving) target amidst obstacles}

Fig. \ref{fig:experiment_spirit} shows Spirit tracking a human in a previously unexplored environment, cluttered with both catalogued obstacles (whose number and placement is unknown in advance) as well as completely unknown obstacles. The robot uses familiar obstacles to both localize itself against them \cite{Bowman2017} and reactively navigate around them. Fig. \ref{fig:environments} summarizes the wide diversity of \`a-priori unexplored environments, with different lighting conditions, successfully navigated indoors (by Turtlebot and Spirit) and outdoors (by Spirit), while tracking humans\footnote{Collision avoidance when the robot gets close to the tracked human is guaranteed with the use of the onboard LIDAR; the human is treated as an unknown obstacle and the robot tries to keep separation and avoid collision (with formal guarantees assuming the conditions of Definition \ref{definition:non_adversarial}).} along thousands of body lengths.

Note that the formal results of Section \ref{sec:reactiveplanning} require that unknown obstacles be convex. However, here we clutter the environment with a mix of unknown obstacles -- some convex, but others of more complicated non-convex shapes (e.g., unknown walls) -- to establish empirical robustness in urban environments that are out of scope of the underlying theory. In such settings, that move beyond the formal assumptions outlined in Section \ref{sec:problemformulation}, the robot might converge to undesired local minima behind non-convex obstacles from a subset of (unfavorable) initial conditions (see Fig. \ref{fig:rrt_narrow_passage}-b); however, collision avoidance is still guaranteed by the onboard LIDAR.

As anticipated, the few failures we recorded were associated with the inability of the SLAM algorithm to localize the robot in long, featureless environments. However, it should be noted that even when the robot or object localization process fails, collision avoidance is still guaranteed with the use of the onboard LIDAR. Nevertheless, collisions could result with obstacles that cannot be detected by the 2D horizontal LIDAR (e.g., the red gascan shown in Fig. \ref{fig:experiment_human_stop}). One could still think of extensions to the presented sensory infrastructure (e.g., the use of a 3D LIDAR) that could at least still guarantee safety under such circumstances.

\subsubsection{Logical reaction using predefined semantics}

In the second set of experimental runs, we exploit the new online semantic capabilities to introduce logic in our reactive tracking process. For example, Fig. \ref{fig:experiment_human_stop} depicts a tracking task requiring the robot to respond to the human's stop signal (raised left or right hand) by returning to its starting position. The supplementary video presents several other semantically specified tasks requiring autonomous reactions of both a logical as well as geometric nature (all involving negotiation of novel environments from the arbitrary geometric circumstances associated with different contexts of logical triggers). 


\section{CONCLUSION AND FUTURE WORK}
\label{sec:conclusion}

This paper presents a reactive planner that can provably safely semantically engage non-adversarial moving targets in planar workspaces, cluttered with an arbitrary mix of catalogued obstacles, using both a wheeled robot and a dynamic legged platform. Future work seeks to extend past hierarchical mobile manipulation schemes using early versions of this architecture \cite{Vasilopoulos_Topping_Vega-Brown_Roy_Koditschek_2018} to incorporate both more dexterous manipulation \cite{topping_vasilopoulos_de_koditschek_2019} as well as logically complex abstract specification (e.g., using temporal logic \cite{kress-gazit-2009}). In the longer term, we believe that concepts from the literature on convex decomposition of polyhedra \cite{lien-amato-2007} may afford a generalization beyond our present restriction to 2D workspaces toward the challenge of navigating partially known environments in higher dimension.






\bibliographystyle{IEEEtran}
\bibliography{IEEEabrv,references}

\begin{thebibliography}{10}
\providecommand{\url}[1]{#1}
\csname url@rmstyle\endcsname
\providecommand{\newblock}{\relax}
\providecommand{\bibinfo}[2]{#2}
\providecommand\BIBentrySTDinterwordspacing{\spaceskip=0pt\relax}
\providecommand\BIBentryALTinterwordstretchfactor{4}
\providecommand\BIBentryALTinterwordspacing{\spaceskip=\fontdimen2\font plus
\BIBentryALTinterwordstretchfactor\fontdimen3\font minus
  \fontdimen4\font\relax}
\providecommand\BIBforeignlanguage[2]{{%
\expandafter\ifx\csname l@#1\endcsname\relax
\typeout{** WARNING: IEEEtran.bst: No hyphenation pattern has been}%
\typeout{** loaded for the language `#1'. Using the pattern for}%
\typeout{** the default language instead.}%
\else
\language=\csname l@#1\endcsname
\fi
#2}}

\bibitem{farber_topology}
M.~Farber, ``{Topology of Robot Motion Planning},'' \emph{Morse Theor. Methods
  in Nonl. Anal. and in Sympl. Topology}, pp. 185--230, 2006.

\bibitem{rimon1992}
E.~Rimon and D.~E. Koditschek, ``{Exact Robot Navigation Using Artificial
  Potential Functions},'' \emph{{IEEE Trans. Robotics and Automation}}, vol.~8,
  no.~5, pp. 501--518, 1992.

\bibitem{Paternain_Koditschek_Ribeiro_2017}
S.~Paternain, D.~E. Koditschek, and A.~Ribeiro, ``{Navigation Functions for
  Convex Potentials in a Space with Convex Obstacles},'' \emph{IEEE Trans.
  Automatic Control}, 2017.

\bibitem{Ilhan_Johnson_Koditschek_2018}
B.~D. Ilhan, A.~M. Johnson, and D.~E. Koditschek, ``Autonomous legged hill
  ascent,'' \emph{J. Field Robotics}, vol. 35 (5), pp. 802--832, 2018.

\bibitem{Arslan_Koditschek_2018}
O.~Arslan and D.~E. Koditschek, ``{Sensor-Based Reactive Navigation in Unknown
  Convex Sphere Worlds},'' \emph{{Int. J. Robotics Research}}, vol.~38, no.
  1-2, pp. 196--223, Jul 2018.

\bibitem{karaman_frazzoli_ICRA2012}
S.~Karaman and E.~Frazzoli, ``High-speed flight in an ergodic forest,'' in
  \emph{{IEEE Int. Conf. Robotics and Automation}}, 2012, pp. 2899--2906.

\bibitem{noreen-khan-habib-2016}
I.~Noreen, A.~Khan, and Z.~Habib, ``{Optimal path planning using RRT* based
  approaches: a survey and future directions},'' \emph{{Int. J. Advanced
  Computer Science and Applications}}, vol.~7, no.~11, pp. 97--107, 2016.

\bibitem{ghostspirit}
\BIBentryALTinterwordspacing
{Ghost Robotics}, ``{Spirit 40}.'' [Online]. Available:
  \url{http://ghostrobotics.io}
\BIBentrySTDinterwordspacing

\bibitem{Pavlakos2017}
G.~Pavlakos, X.~Zhou, A.~Chan, K.~G. Derpanis, and K.~Daniilidis, ``{6-DoF
  object pose from semantic keypoints},'' in \emph{IEEE Int. Conf. Robotics and
  Automation}, May 2017, pp. 2011--2018.

\bibitem{Bowman2017}
S.~L. Bowman, N.~Atanasov, K.~Daniilidis, and G.~J. Pappas, ``{Probabilistic
  data association for semantic SLAM},'' in \emph{IEEE Int. Conf. Robotics and
  Automation}, May 2017, pp. 1722--1729.

\bibitem{vasilopoulos_pavlakos_schmeckpeper_daniilidis_koditschek_2019}
V.~Vasilopoulos \emph{et~al.}, ``{Reactive Navigation in Partially Familiar
  Planar Environments Using Semantic Perceptual Feedback},'' \emph{{Under
  review, arXiv: 2002.08946}}, 2020.

\bibitem{Gupta_2017}
S.~Gupta, J.~Davidson, S.~Levine, R.~Sukthankar, and J.~Malik, ``{Cognitive
  Mapping and Planning for Visual Navigation},'' in \emph{{IEEE CVPR}}, 2017,
  pp. 7272--7281.

\bibitem{savinov-dosovitskiy-koltun-2018}
N.~Savinov, A.~Dosovitskiy, and V.~Koltun, ``{Semi-parametric topological
  memory for navigation},'' \emph{{arXiv: 1803.00653}}, 2018.

\bibitem{shen-xu-zhu--guibas-feifei-savarese-2019}
W.~B. Shen, D.~Xu, Y.~Zhu, L.~J. Guibas, L.~Fei-Fei, and S.~Savarese,
  ``{Situational Fusion of Visual Representation for Visual Navigation},'' in
  \emph{IEEE Int. Conf. Computer Vision}, 2019, pp. 2881--2890.

\bibitem{meng-ratliff-xiang-fox-2019}
X.~Meng, N.~Ratliff, Y.~Xiang, and D.~Fox, ``{Neural Autonomous Navigation with
  Riemannian Motion Policy},'' in \emph{IEEE Int. Conf. Robotics and
  Automation}, 2019, pp. 8860--8866.

\bibitem{janson-hu-pavone-2018}
L.~Janson, T.~Hu, and M.~Pavone, ``{Safe Motion Planning in Unknown
  Environments: Optimality Benchmarks and Tractable Policies},'' in
  \emph{Robotics: Science and Systems}, 2018.

\bibitem{lahijanian-2016}
M.~Lahijanian \emph{et~al.}, ``{Iterative Temporal Planning in Uncertain
  Environments With Partial Satisfaction Guarantees},'' \emph{{IEEE Trans.
  Robotics}}, vol.~32, no.~3, pp. 583--599, 2016.

\bibitem{bajcsy2019efficient}
A.~Bajcsy, S.~Bansal, E.~Bronstein, V.~Tolani, and C.~J. Tomlin, ``{An
  Efficient Reachability-Based Framework for Provably Safe Autonomous
  Navigation in Unknown Environments},'' \emph{arXiv: 1905.00532}, 2019.

\bibitem{kousik-vaskov-bu-johnson-vasudevan-2018}
S.~Kousik \emph{et~al.}, ``{Bridging the Gap Between Safety and Real-Time
  Performance in Receding-Horizon Trajectory Design for Mobile Robots},''
  \emph{{arXiv: 1809.06746}}, 2018.

\bibitem{vasilopoulos_koditschek_WAFR2018}
V.~Vasilopoulos and D.~E. Koditschek, ``{Reactive Navigation in Partially Known
  Non-Convex Environments},'' in \emph{13th Int. Workshop on the Algorithmic
  Foundations of Robotics (WAFR)}, 2018.

\bibitem{kolotouros2019learning}
N.~Kolotouros, G.~Pavlakos, M.~J. Black, and K.~Daniilidis, ``Learning to
  reconstruct 3{D} human pose and shape via model-fitting in the loop,'' in
  \emph{IEEE Int. Conf. Computer Vision}, 2019, pp. 2252--2261.

\bibitem{greene-1983}
D.~H. Greene, ``{The decomposition of polygons into convex parts},''
  \emph{{Computational Geometry}}, vol.~1, pp. 235--259, 1983.

\bibitem{otte-2015}
M.~Otte and E.~Frazzoli, ``{RRT$^{X}$: Asymptotically optimal single-query
  sampling-based motion planning with quick replanning},'' \emph{{Int. J. of
  Robotics Research}}, vol.~35, no.~7, pp. 797--822, 2015.

\bibitem{shapiro2007}
V.~Shapiro, ``{Semi-analytic geometry with R-functions},'' \emph{{Acta
  Numerica}}, vol.~16, pp. 239--303, 2007.

\bibitem{cgal}
\BIBentryALTinterwordspacing
{The CGAL Project}, \emph{{CGAL} User and Reference Manual}, {4.14}~ed.\hskip
  1em plus 0.5em minus 0.4em\relax {CGAL Editorial Board}, 2019. [Online].
  Available: \url{https://doc.cgal.org}
\BIBentrySTDinterwordspacing

\bibitem{lien-amato-2004}
J.-M. Lien and N.~M. Amato, ``{Approximate convex decomposition of polygons},''
  \emph{{Comp. Geometry}}, vol.~35, no. 1-2, pp. 100--123, 2006.

\bibitem{keil-snoeyink-2002}
M.~Keil and J.~Snoeyink, ``{On the time bound for convex decomposition of
  simple polygons},'' \emph{{Int. J. Computational Geometry \& Applications}},
  vol.~12, no.~3, pp. 181--192, 2002.

\bibitem{keil-convex-decomposition}
J.~M. Keil, ``{Decomposing a Polygon into Simpler Components},'' \emph{{SIAM
  Journal on Computing}}, vol.~14, no.~4, pp. 799--817, 1985.

\bibitem{yolov3}
J.~Redmon and A.~Farhadi, ``Yolov3: An incremental improvement,'' \emph{arXiv:
  1804.02767}, 2018.

\bibitem{arslan_kod_ICRA2017}
O.~Arslan and D.~E. Koditschek, ``{Smooth Extensions of Feedback Motion
  Planners via Reference Governors},'' in \emph{IEEE Int. Conf. Robotics and
  Automation}, 2017, pp. 4414--4421.

\bibitem{hsu-2000}
J.~C. Latombe and D.~Hsu, ``Randomized single-query motion planning in
  expansive spaces,'' Ph.D. dissertation, Stanford University, 2000.

\bibitem{turtlebot}
\BIBentryALTinterwordspacing
{TurtleBot2}, ``{Open-source robot development kit for apps on wheels},'' 2019.
  [Online]. Available: \url{https://www.turtlebot.com/turtlebot2/}
\BIBentrySTDinterwordspacing

\bibitem{boost}
B.~Schling, \emph{{The Boost C++ Libraries}}.\hskip 1em plus 0.5em minus
  0.4em\relax {XML Press}, 2011.

\bibitem{Vasilopoulos_Topping_Vega-Brown_Roy_Koditschek_2018}
V.~Vasilopoulos \emph{et~al.}, ``{Sensor-Based Reactive Execution of Symbolic
  Rearrangement Plans by a Legged Mobile Manipulator},'' in \emph{IEEE/RSJ Int.
  Conf. Intelligent Robots and Systems}, 2018, pp. 3298--3305.

\bibitem{topping_vasilopoulos_de_koditschek_2019}
T.~T. Topping, V.~Vasilopoulos, A.~De, and D.~E. Koditschek, ``Composition of
  templates for transitional pedipulation behaviors,'' in \emph{Int. Symposium
  on Robotics Research}, 2019.

\bibitem{kress-gazit-2009}
H.~Kress-Gazit, G.~E. Fainekos, and G.~J. Pappas, ``{Temporal-logic-based
  reactive mission and motion planning},'' \emph{{IEEE Trans. Robotics}},
  vol.~25, no.~6, pp. 1370--1381, 2009.

\bibitem{lien-amato-2007}
J.-M. Lien and N.~M. Amato, ``{Approximate convex decomposition of
  polyhedra},'' in \emph{ACM Symp. Solid and Phys. Model.}, 2007, pp. 121--131.

\bibitem{hirsch_1976}
M.~W. Hirsch, \emph{{Differential Topology}}.\hskip 1em plus 0.5em minus
  0.4em\relax Springer, 1976.

\bibitem{massey1992}
W.~S. Massey, ``{Sufficient conditions for a local homeomorphism to be
  injective},'' \emph{{Topology and its Applications}}, vol.~47, pp. 133--148,
  1992.

\bibitem{Rimon_Koditschek_1989}
E.~Rimon and D.~E. Koditschek, ``{The Construction of Analytic Diffeomorphisms
  for Exact Robot Navigation on Star Worlds},'' \emph{Trans. American
  Mathematical Society}, vol. 327, no.~1, pp. 71--116, 1989.

\end{thebibliography}

\appendices
\section{Controller for Differential Drive Robots}
\label{appendix:differential_drive}

Since a differential drive robot, whose dynamics are given by\footnote{We use the ordered set notation $(*,*,\ldots)$ and the matrix notation $\begin{bmatrix} * & * & \ldots \end{bmatrix}^\top$ for vectors interchangeably.} $\dot{\robotpositionunicycle} = \mathbf{B}(\robotorientation) \controlunicycle \label{eq:unicycle_dynamics}$,
with $\mathbf{B}(\robotorientation):=\begin{bmatrix} \cos\robotorientation & \sin\robotorientation & 0 \\ 0 & 0 & 1 \end{bmatrix}^\top$ and $\controlunicycle:=(\linearinput,\angularinput)$, with $\linearinput,\angularinput \in \mathbb{R}$ the linear and angular input respectively, operates in $SE(2)$ instead of $\mathbb{R}^2$, we first need a smooth diffeomorphism $\diffeounicycle:\freespacemapped \times S^1 \rightarrow \hybridfreespacemodemodel \times S^1$ away from sharp corners on the boundary of $\freespacemapped \times S^1$, and then establish the results about our controller.

Following our previous work \cite{vasilopoulos_koditschek_WAFR2018,vasilopoulos_pavlakos_schmeckpeper_daniilidis_koditschek_2019}, we construct our map $\diffeounicycle$ from $\freespacemapped \times S^1$ to $\hybridfreespacemodemodel \times S^1$ as $\robotpositionunicyclemodel = (\robotpositionmodel,\robotorientationmodel) = \diffeounicycle(\robotpositionunicycle):=(\diffeo(\robotposition),\angletransform(\robotpositionunicycle)) \label{eq:diffeo_unicycle}$, with $\robotpositionunicycle=(\robotposition,\robotorientation) \in \freespacemapped \times S^1$, $\robotpositionunicyclemodel:=(\robotpositionmodel,\robotorientationmodel) \in \hybridfreespacemodemodel \times S^1$ and $ \robotorientationmodel=\angletransform(\robotpositionunicycle) := \angle(\directionvector(\robotpositionunicycle)) \label{eq:phi}$. Here, $\angle \directionvector:=\text{atan2}(e_2,e_1)$ and $\directionvector(\robotpositionunicycle) = \mathrm{\Pi}_\robotpositionmodel \cdot D_{\robotpositionunicycle} \diffeounicycle \cdot \mathbf{B}(\robotorientation) \cdot \begin{bmatrix} 1 \\ 0 \end{bmatrix} = D_\robotposition \diffeo \begin{bmatrix} \cos\robotorientation \\ \sin\robotorientation \end{bmatrix}$, with $\mathrm{\Pi}_\robotpositionmodel$ denoting the projection onto the first two components. We show in \cite{vasilopoulos_pavlakos_schmeckpeper_daniilidis_koditschek_2019} that $\diffeounicycle$ is a $C^\infty$ diffeomorphism from $\freespacemapped \times S^1$ to $\freespacemodel \times S^1$ away from sharp corners, none of which lie in the interior of $\freespacemapped \times S^1$.

Based on the above, we can then write $\dot{\robotpositionunicyclemodel} = \begin{bmatrix} \dot{\robotpositionmodel} \\ \dot{\robotorientationmodel} \end{bmatrix} = \frac{d}{dt} \begin{bmatrix} \diffeo(\robotposition) \\ \angletransform(\robotpositionunicycle) \end{bmatrix} = \mathbf{B}(\robotorientationmodel) \controlunicyclemodel^\hybridmode \label{eq:unicycle_dynamics_se2}$
with $\controlunicyclemodel^\hybridmode = (\linearinputmodel^\hybridmode,\angularinputmodel^\hybridmode)$, and the inputs $(\linearinputmodel^\hybridmode,\angularinputmodel^\hybridmode)$ related to $(\linearinput^\hybridmode,\angularinput^\hybridmode)$ through $\linearinputmodel^\hybridmode = ||\directionvector(\robotpositionunicycle)|| \, \linearinput^\hybridmode$ and $\angularinputmodel^\hybridmode = \linearinput^\hybridmode D_\robotposition\angletransform \begin{bmatrix} \cos\robotorientation \\ \sin\robotorientation \end{bmatrix} + \dfrac{\partial \angletransform}{\partial \robotorientation} \angularinput^\hybridmode$, with $D_\robotposition\angletransform = \begin{bmatrix}
\frac{\partial \angletransform}{\partial x} & \frac{\partial \angletransform}{\partial y}
\end{bmatrix}$. The idea now is to use the control strategy in \cite{Arslan_Koditschek_2018} to find inputs $\linearinputmodel^\hybridmode,\angularinputmodel^\hybridmode$ in $\freespacemodel \times S^1$, and then use the relations above to find the actual inputs $\linearinput^\hybridmode,\angularinput^\hybridmode$ in $\freespacemapped \times S^1$ that achieve $\linearinputmodel^\hybridmode,\angularinputmodel^\hybridmode$ as
\begin{subequations} \label{eq:control_unicycle}
\begin{eqnarray}
& \linearinput^\hybridmode =\dfrac{k_v \, \linearinputmodel^\hybridmode}{||\directionvector(\robotpositionunicycle)||} \\
& \angularinput^\hybridmode = \left(\dfrac{\partial \angletransform}{\partial \robotorientation}\right)^{-1} \left(k_\omega \, \angularinputmodel^\hybridmode-\linearinput^\hybridmode D_\robotposition\angletransform \begin{bmatrix}
\cos\robotorientation \\ \sin\robotorientation
\end{bmatrix} \right)
\end{eqnarray}
\end{subequations}
with $k_v,k_\omega>0$ fixed gains.

\section{Proofs of Main Results}
\label{appendix:proofs}

\begin{proof}[Proof of Proposition \ref{proposition:diffeo_purging}]
We follow similar patterns to the proof of \cite[Proposition 1]{vasilopoulos_pavlakos_schmeckpeper_daniilidis_koditschek_2019}. We first need to show that the functions $\switch{j_i}, \deformingfactor{j_i} : \freespacemappedpurging{j_i} \rightarrow \mathbb{R}$ are smooth away from the polygon vertices, none of which lies in the interior of $\freespacemappedpurging{j_i}$. We begin with $\switch{j_i}$. First of all, with the procedure outlined in \cite{vasilopoulos_koditschek_WAFR2018}, the only points where $\innerpolygonimplicit{j_i}$ and $\outerpolygonimplicit{j_i}$ are not smooth are vertices of $\innerpolygon{j_i}$ and $\outerpolygon{j_i}$ respectively. We use the $C^\infty$ function $\zeta_\mu:\mathbb{R} \rightarrow \mathbb{R}$ \cite{hirsch_1976} described by
\begin{equation}
\label{eq:zeta}
\zeta_\mu(\chi) = \left\{ \begin{matrix}
e^{-\mu/\chi}, & \quad \chi>0 \\
0,  & \quad \chi \leq 0
\end{matrix}\right.
\end{equation}
and parametrized by $\mu > 0$, that has derivative
\begin{equation} \label{eq:zeta_derivative}
\zeta_\mu'(\chi) = \left\{ \begin{matrix}
\frac{\mu \, \zeta_\mu(\chi)}{\chi^{2}}, & \quad \chi>0 \\
0,  & \quad \chi \leq 0
\end{matrix}\right.
\end{equation}
and define the smooth auxiliary $C^\infty$ switches
\begin{align}
    \innerpolygonsigma{j_i}(\robotposition) & := \eta_{\innerpolygontune{j_i},\innerpolygondistance{j_i}} \circ \innerpolygonimplicit{j_i}(\robotposition) \label{eq:sigma_gamma_ji} \\ 
    \outerpolygonsigma{j_i}(\robotposition) & := \zeta_{\outerpolygontune{j_i}} \circ \frac{\outerpolygonimplicit{j_i}(\robotposition)}{||\robotposition-\diffeocenter{j_i}||} \label{eq:sigma_delta_ji}
\end{align}
with $\eta_{\mu,\epsilon}(\chi) := \zeta_\mu(\epsilon - \chi)/\zeta_\mu(\epsilon)$, and $\innerpolygontune{j_i}, \outerpolygontune{j_i}, \innerpolygondistance{j_i} > 0$ tunable parameters. We note that $\outerpolygonsigma{j_i}$ is smooth everywhere, since $\diffeocenter{j_i}$ does not belong in $\freespacemappedpurging{j_i}$ and $\outerpolygonimplicit{j_i}$ is exactly 0 on the vertices of $\outerpolygon{j_i}$. Therefore, by defining $\switch{j_i}$ as 
\begin{equation}
    \switch{j_i}(\robotposition):= \left\{ \begin{matrix} \frac{\innerpolygonsigma{j_i}(\robotposition)\outerpolygonsigma{j_i}(\robotposition)}{\innerpolygonsigma{j_i}(\robotposition)\outerpolygonsigma{j_i}(\robotposition) + \left(1-\innerpolygonsigma{j_i}(\robotposition)\right)}, & \robotposition \neq \robotposition_{1j_i}, \robotposition_{2j_i} \\ 1, & \robotposition = \robotposition_{1j_i},\robotposition_{2j_i} \end{matrix} \right. \label{eq:sigma_ji}
\end{equation}
with $\robotposition_{1j_i}\robotposition_{2j_i}$ defining the shared hyperplane between $j_i$ and $p(j_i)$, we get that $\switch{j_i}$ can only be non-smooth on the vertices of $\innerpolygon{j_i}$ except for $\diffeocenter{j_i}$ (i.e., on the vertices of the polygon $j_i$), and on points where its denominator becomes zero. Since both $\innerpolygonsigma{j_i}$ and $\outerpolygonsigma{j_i}$ vary between 0 and 1, this can only happen when $\innerpolygonsigma{j_i} (\robotposition) = 1$ and $\outerpolygonsigma{j_i}(\robotposition) = 0$, i.e., only on $\robotposition_{1j_i}$ and $\robotposition_{2j_i}$. The fact that $\switch{j_i}$ is smooth everywhere else derives immediately from the fact that $\outerpolygonsigma{j_i}$ is a smooth function, and $\innerpolygonsigma{j_i}$ is smooth everywhere except for the polygon vertices. 

On the other hand, the singular points of the deforming factor $\deformingfactor{j_i}$, defined as
\begin{equation}
    \deformingfactor{j_i}(\robotposition):= \frac{\left(\robotposition_{1j_i} - \diffeocenter{j_i} \right)^\top \sharednormal{j_i}}{\left(\robotposition - \diffeocenter{j_i} \right)^\top \sharednormal{j_i}} \label{eq:deforming_factor_purging}
\end{equation}
with
\begin{equation}
    \sharednormal{j_i} := \mathbf{R}_{\frac{\pi}{2}} \frac{\robotposition_{2j_i}-\robotposition_{1j_i}}{||\robotposition_{2j_i}-\robotposition_{1j_i}||}, \quad \mathbf{R}_{\frac{\pi}{2}}:= \begin{bmatrix} 0 & -1 \\ 1 & 0 \end{bmatrix}
\end{equation}
the normal vector corresponding to the shared edge between $j_i$ and $p(j_i)$, are the solutions of the equation $(\robotposition-\diffeocenter{j_i})^\top \sharednormal{j_i} = 0$, which lie on the hyperplane passing through $\diffeocenter{j_i}$ with normal vector $\sharednormal{j_i}$ and, due to the construction of $\outerpolygon{j_i}$ as in Definition \ref{definition:collars}, lie outside of $\outerpolygon{j_i}$ and do not affect the map $\freespacemappedpurging{j_i}$. Hence, the map $\diffeopurging{j_i}$ is smooth everywhere in $\freespacemappedpurging{j_i}$, except for the vertices of the polygon $j_i$, as a composition of smooth functions with the same properties.

Now, in order to prove that $\diffeopurging{j_i}$ is a $C^\infty$ diffeomorphism away from the vertices of $j_i$, we follow the procedure outlined in \cite{massey1992}, also followed in \cite{Rimon_Koditschek_1989}, to show that
\begin{enumerate}
    \item $\diffeopurging{j_i}$ has a non-singular differential on $\freespacemappedpurging{j_i}$ except for the vertices of polygon $j_i$.
    \item $\diffeopurging{j_i}$ preserves boundaries, i.e., $\diffeopurging{j_i}(\partial_k\freespacemappedpurging{j_i}) \subset \partial_k\freespacemappedpurging{p(j_i)}$, with $k$ spanning both the indices of familiar obstacles $\knownobstaclesetdilatedmappeddiskindex$, $\knownobstaclesetdilatedmappedintrusionindex$ as well as the indices of unknown obstacles $\unknownobstaclesetdilatedsemanticindex$, and $\partial_k \freespace$ the $k$-th connected component of the boundary of $\freespace$ with $\partial_0\freespace$ the outer boundary of $\freespace$.
    \item the boundary components of $\freespacemappedpurging{j_i}$ and $\freespacemappedpurging{p(j_i)}$ are pairwise homeomorphic, i.e., $\partial_k\freespacemappedpurging{j_i} \cong \partial_k \freespacemappedpurging{p(j_i)}$, with $k$ spanning both the indices of familiar obstacles $\knownobstaclesetdilatedmappeddiskindex$, $\knownobstaclesetdilatedmappedintrusionindex$ as well as the indices of unknown obstacles $\unknownobstaclesetdilatedsemanticindex$.
\end{enumerate}
We begin with Property 1 and examine the space away from the vertices of $j_i$. The case where $\outerpolygonsigma{j_i}$ is 0 (outside of the polygonal collar $\outerpolygon{j_i}$) is not interesting, since $\diffeopurging{j_i}$ defaults to the identity map and $D_\robotposition\diffeopurging{j_i} = \mathbf{I}$. When $\outerpolygonsigma{j_i}$ is not 0, we can compute the jacobian of the map as
\begin{align}
    D_\robotposition\diffeopurging{j_i} = & \left(\deformingfactor{j_i}(\robotposition)-1\right) (\robotposition-\diffeocenter{j_i}) \nabla \switch{j_i}(\robotposition)^\top \nonumber \\
    & + \switch{j_i}(\robotposition)(\robotposition-\diffeocenter{j_i}) \nabla \deformingfactor{j_i}(\robotposition)^\top \nonumber \\
    & + \left[ 1 + \switch{j_i}(\robotposition)\left(\deformingfactor{j_i}(\robotposition)-1\right) \right] \mathbf{I} \label{eq:jacobian_purging}
\end{align}
For the deforming factor $\deformingfactor{j_i}$ we compute from \eqref{eq:deforming_factor_purging}
\begin{equation}
    \nabla \deformingfactor{j_i}(\robotposition) = -\frac{\left(\robotposition_{1j_i} - \diffeocenter{j_i} \right) ^\top \sharednormal{j_i}}{\left[\left(\robotposition - \diffeocenter{j_i} \right) ^\top \sharednormal{j_i}\right]^2} \sharednormal{j_i}
\end{equation}
Note that we interestingly get
\begin{equation}
    \left(\robotposition-\diffeocenter{j_i} \right)^\top \nabla \deformingfactor{j_i}(\robotposition) = -\deformingfactor{j_i}(\robotposition) \label{eq:nu_inner_purging}
\end{equation}
From \eqref{eq:jacobian_purging} it can be seen that $D_\robotposition\diffeopurging{j_i} = \mathbf{A} + \mathbf{u}\mathbf{v}^\top$ with $\mathbf{A} = \left[ 1 + \switch{j_i}(\robotposition)\left(\deformingfactor{j_i}(\robotposition)-1\right) \right] \mathbf{I}$, $\mathbf{u} = \robotposition-\diffeocenter{j_i}$ and $\mathbf{v} = \left(\deformingfactor{j_i}(\robotposition)-1\right)\nabla \switch{j_i}(\robotposition) + \switch{j_i}(\robotposition) \nabla \deformingfactor{j_i}(\robotposition)$.

Due to the fact that $0 \leq \switch{j_i}(\robotposition) \leq 1$ and $0 < \deformingfactor{j_i}(\robotposition) < 1$ in the interior of an admissible polygonal collar $\outerpolygon{j_i}$ (see Definition \ref{definition:collars}), we get $1 + \switch{j_i}(\robotposition)\left(\deformingfactor{j_i}(\robotposition)-1\right) > 0$. Hence, $\mathbf{A}$ is invertible, and by using the matrix determinant lemma and \eqref{eq:nu_inner_purging}, the determinant of $D_\robotposition\diffeopurging{j_i}$ can be computed as
\begin{align}
    & \text{det}(D_\robotposition\diffeopurging{j_i}) = \text{det}\mathbf{A} + (\text{det}\mathbf{A})\mathbf{v}^\top \mathbf{A}^{-1} \mathbf{u} \nonumber \\
    & = \left[ 1 + \switch{j_i}(\robotposition)\left(\deformingfactor{j_i}(\robotposition)-1\right) \right] \cdot \nonumber \\
    & \cdot \left[ \left(1-\switch{j_i}(\robotposition)\right) + \left(\deformingfactor{j_i}(\robotposition)-1\right)(\robotposition-\diffeocenter{j_i})^\top \nabla \switch{j_i}(\robotposition) \right]
\end{align}
Similarly the trace of $D_\robotposition\diffeopurging{j_i}$ can be computed as
\begin{align}
    & \text{tr}(D_\robotposition\diffeopurging{j_i}) = \left[ 1 + \switch{j_i}(\robotposition)\left(\deformingfactor{j_i}(\robotposition)-1\right) \right] + \left(1-\switch{j_i}(\robotposition)\right) \nonumber \\
    & + \left(\deformingfactor{j_i}(\robotposition)-1\right)(\robotposition-\diffeocenter{j_i})^\top \nabla \switch{j_i}(\robotposition)
\end{align}

Also, by construction of the switch $\switch{j_i}$, we see that $\nabla \switch{j_i}(\robotposition) = \mathbf{0}$ when $\switch{j_i}(\robotposition) = 0$. Hence, using the above expressions, we can show that $\text{det}(D_\robotposition\diffeopurging{j_i}), \text{tr}(D_\robotposition\diffeopurging{j_i}) > 0$ (and therefore establish that $D_\robotposition\diffeopurging{j_i}$ is not singular in the interior of $\outerpolygon{j_i}$, since $\freespacemappedpurging{j_i} \subseteq \mathbb{R}^2$) by showing that $(\robotposition-\diffeocenter{j_i})^\top \nabla \switch{j_i}(\robotposition) < 0$ when $\switch{j_i}(\robotposition) > 0$, where
\begin{align}
    \nabla \switch{j_i}(\robotposition) = & \frac{\outerpolygonsigma{j_i}(\robotposition) }{\left[ \innerpolygonsigma{j_i}(\robotposition) \outerpolygonsigma{j_i}(\robotposition) + \left(1- \innerpolygonsigma{j_i}(\robotposition) \right) \right]^2} \nabla \innerpolygonsigma{j_i}(\robotposition) \nonumber \\
    & + \frac{\innerpolygonsigma{j_i}(\robotposition) \left(1-\innerpolygonsigma{j_i}(\robotposition) \right)}{\left[ \innerpolygonsigma{j_i}(\robotposition) \outerpolygonsigma{j_i}(\robotposition) + \left(1- \innerpolygonsigma{j_i}(\robotposition) \right) \right]^2} \nabla \outerpolygonsigma{j_i}(\robotposition)
\end{align}
with
\begin{equation}
    \nabla \innerpolygonsigma{j_i}(\robotposition) = \left\{ \begin{matrix} -\dfrac{\innerpolygontune{j_i} \innerpolygonsigma{j_i}(\robotposition)}{\left(\innerpolygondistance{j_i} - \innerpolygonimplicit{j_i}(\robotposition) \right)^2} \nabla \innerpolygonimplicit{j_i}(\robotposition), & \innerpolygonimplicit{j_i}(\robotposition) < \innerpolygondistance{j_i} \\
    \mathbf{0}, & \innerpolygonimplicit{j_i}(\robotposition) \geq \innerpolygondistance{j_i} \end{matrix} \right.
\end{equation}
\begin{equation}
    \nabla \outerpolygonsigma{j_i}(\robotposition) = \left\{ \begin{matrix} \dfrac{\outerpolygontune{j_i} \outerpolygonsigma{j_i}(\robotposition)}{\alpha_{j_i}(\robotposition)^2} \nabla \alpha_{j_i}(\robotposition), & \outerpolygonimplicit{j_i}(\robotposition) > 0 \\
    \mathbf{0}, & \outerpolygonimplicit{j_i}(\robotposition) \leq 0 \end{matrix} \right.
\end{equation}
and $\alpha_{j_i}(\robotposition) := \outerpolygonimplicit{j_i}(\robotposition)/||\robotposition-\diffeocenter{j_i}||$. Therefore, it suffices to show that when $\switch{j_i}(\robotposition) > 0$:
\begin{align}
    (\robotposition-\diffeocenter{j_i})^\top \nabla \innerpolygonimplicit{j_i}(\robotposition) > 0 \label{eq:condition_gamma} \\
    (\robotposition-\diffeocenter{j_i})^\top \nabla \alpha_{j_i}(\robotposition) < 0 \label{eq:condition_delta}
\end{align}

Following the procedure outlined in \cite{vasilopoulos_koditschek_WAFR2018} for the implicit representation of polygonal obstacles and assuming that the polygon $\innerpolygon{j_i}$ has $m$ sides, we can describe $\innerpolygon{j_i}$ with the implicit function $\innerpolygonimplicit{j_i} = \neg \left( (\innerpolygonimplicit{1j_i}\wedge \innerpolygonimplicit{2j_i}) \wedge \ldots \wedge \innerpolygonimplicit{mj_i} \right)$, with the companion R-function \cite{shapiro2007} of the logic negation for a function $x$ defined as $\neg x := -x$, the companion R-function of the logic conjunction $\wedge$ for two functions $x_1, x_2$ defined as $x_1 \wedge x_2 := x_1+x_2-\left(x_1^p+x_2^p\right)^\frac{1}{p}$, and $\innerpolygonimplicit{kj_i}$ the $k$-th hyperplane equation describing $\innerpolygon{j_i}$, given as $\innerpolygonimplicit{kj_i}(\robotposition) := (\robotposition - \robotposition_{kj_i}) ^\top \sharednormal{kj_i}$. Note here that the first two hyperplanes $\innerpolygonimplicit{1j_i}$ and $\innerpolygonimplicit{2j_i}$ pass through the center $\diffeocenter{j_i}$, i.e., we can write $\innerpolygonimplicit{1j_i}(\robotposition) = (\robotposition-\diffeocenter{j_i}) ^\top \sharednormal{1j_i}$ and $\innerpolygonimplicit{2j_i}(\robotposition) = (\robotposition-\diffeocenter{j_i}) ^\top \sharednormal{2j_i}$. Based on this observation, it is easy to derive the following expression for any $\robotposition$ that satisfies $\switch{j_i}(\robotposition) > 0$
\begin{equation}
    (\robotposition-\diffeocenter{j_i}) ^\top \nabla (\innerpolygonimplicit{1j_i} \wedge \innerpolygonimplicit{2j_i}) = \innerpolygonimplicit{1j_i} \wedge \innerpolygonimplicit{2j_i}
\end{equation}
We can then similarly compute
\begin{align}
    & \nabla \left(\left(\innerpolygonimplicit{1j_i} \wedge \innerpolygonimplicit{2j_i}\right) \wedge \innerpolygonimplicit{3j_i} \right) = \left(1 - \tfrac{\innerpolygonimplicit{3j_i}}{\sqrt{(\innerpolygonimplicit{1j_i} \wedge \innerpolygonimplicit{2j_i})^2 + \innerpolygonimplicit{3j_i}^2}} \right) \nabla \innerpolygonimplicit{3j_i} \nonumber \\
    & + \left(1 - \tfrac{\innerpolygonimplicit{1j_i} \wedge \innerpolygonimplicit{2j_i}}{\sqrt{(\innerpolygonimplicit{1j_i} \wedge \innerpolygonimplicit{2j_i})^2 + \innerpolygonimplicit{3j_i}^2}} \right) \nabla (\innerpolygonimplicit{1j_i} \wedge \innerpolygonimplicit{2j_i}) \nonumber
\end{align}
and observe that $(\robotposition-\diffeocenter{j_i}) ^\top \nabla \innerpolygonimplicit{3j_i} = (\robotposition-\robotposition_{3j_i}) ^\top \nabla \innerpolygonimplicit{3j_i} - (\diffeocenter{j_i}-\robotposition_{3j_i}) ^\top \nabla \innerpolygonimplicit{3j_i} = \innerpolygonimplicit{3j_i} - (\diffeocenter{j_i}-\robotposition_{3j_i}) ^\top \sharednormal{3j_i} < \innerpolygonimplicit{3j_i}$, which implies that $(\robotposition-\diffeocenter{j_i}) ^\top \nabla \left(\left(\innerpolygonimplicit{1j_i} \wedge \innerpolygonimplicit{2j_i}\right) \wedge \innerpolygonimplicit{3j_i} \right) < \left(\innerpolygonimplicit{1j_i} \wedge \innerpolygonimplicit{2j_i}\right) \wedge \innerpolygonimplicit{3j_i}$.

We can repeat this step inductively for all hyperplanes comprising $\innerpolygon{j_i}$ to show that
\begin{align}
    & (\robotposition-\diffeocenter{j_i}) ^\top \nabla \left( (\innerpolygonimplicit{1j_i}\wedge \innerpolygonimplicit{2j_i}) \wedge \ldots \wedge \innerpolygonimplicit{mj_i} \right) < \nonumber \\
    & \left( (\innerpolygonimplicit{1j_i}\wedge \innerpolygonimplicit{2j_i}) \wedge \ldots \wedge \innerpolygonimplicit{mj_i} \right) \nonumber
\end{align}
The last step is to apply the negation induced by the R-function and arrive at the desired result:
\begin{equation}
    (\robotposition-\diffeocenter{j_i})^\top \nabla \innerpolygonimplicit{j_i}(\robotposition) > \innerpolygonimplicit{j_i}(\robotposition) \geq 0
\end{equation}

The proof of \eqref{eq:condition_delta} follows similar patterns. Here, we focus on $\outerpolygonimplicit{j_i}$. The external polygonal collar $\outerpolygon{j_i}$ can be assumed to have $n$ sides, which means that we can write $\outerpolygonimplicit{j_i} = \left( (\outerpolygonimplicit{1j_i}\wedge \outerpolygonimplicit{2j_i}) \wedge \ldots \wedge \outerpolygonimplicit{nj_i} \right)$. Following the procedure outlined above for the proof of \eqref{eq:condition_gamma}, we can expand each term in the conjunction individually and then combine them to get
\begin{equation}
    (\robotposition-\diffeocenter{j_i}) ^\top \nabla \outerpolygonimplicit{j_i} (\robotposition) < \outerpolygonimplicit{j_i} (\robotposition) \label{eq:delta_inner}
\end{equation}
We also have
\begin{align}
    \nabla \alpha_{j_i}(\robotposition) & = \nabla \left( \frac{\outerpolygonimplicit{j_i}(\robotposition)}{||\robotposition-\diffeocenter{j_i}||} \right) \nonumber \\
    & = \frac{||\robotposition-\diffeocenter{j_i}|| \nabla \outerpolygonimplicit{j_i}(\robotposition) - \outerpolygonimplicit{j_i}(\robotposition)\tfrac{\robotposition-\diffeocenter{j_i}}{||\robotposition-\diffeocenter{j_i}||}}{||\robotposition-\diffeocenter{j_i}||^2}
\end{align}
which gives the desired result using \eqref{eq:delta_inner}
\begin{equation}
    (\robotposition-\diffeocenter{j_i}) \nabla \alpha_{j_i}(\robotposition) = \frac{(\robotposition-\diffeocenter{j_i}) \nabla \outerpolygonimplicit{j_i}(\robotposition) - \outerpolygonimplicit{j_i}(\robotposition)}{||\robotposition-\diffeocenter{j_i}||} < 0
\end{equation}
This concludes the proof that $\diffeopurging{j_i}$ satisfies Property 1.

Next, we focus on Property 2. Pick a point $\robotposition \in \partial_k \freespacemappedpurging{j_i}$. This point could lie:
\begin{enumerate}
    \item on the outer boundary of $\freespacemappedpurging{j_i}$ and away from $\knownobstacledilated_i$
    \item on the boundary of one of the $|\unknownobstaclesetdilatedsemanticindex|$ unknown but visible convex obstacles
    \item on the boundary of one of the $(|\knownobstaclesetdilatedmappeddiskindex|+|\knownobstaclesetdilatedmappedintrusionindex|-1)$ familiar obstacles that are not $\knownobstacledilated_i$
    \item on the boundary of $\knownobstacledilated_i$ but not on the boundary of the polygon $j_i$
    \item on the boundary of the polygon $j_i$
\end{enumerate}
In the first four cases, we have $\diffeopurging{j_i}(\robotposition) = \robotposition$, whereas in the last case, we have
\begin{equation}
    \diffeopurging{j_i}(\robotposition) = \diffeocenter{j_i} + \frac{\left(\robotposition_{1j_i} - \diffeocenter{j_i} \right)^\top \sharednormal{j_i}}{\left(\robotposition - \diffeocenter{j_i} \right)^\top \sharednormal{j_i}} (\robotposition-\diffeocenter{j_i})
\end{equation}
It can be verified that $\left( \diffeopurging{j_i}(\robotposition) - \robotposition_{1j_i} \right)^\top \sharednormal{j_i} = 0$, which means that $\robotposition$ is sent to the shared hyperplane between $j_i$ and $p(j_i)$ as desired. This shows that we always have $\diffeopurging{j_i}(\robotposition) \in \partial_k \freespacemappedpurging{p(j_i)}$ and the map satisfies Property 2. 

Finally, Property 3 derives from above and the fact that each boundary segment $\partial_k \freespacemappedpurging{j_i}$ is an one-dimensional manifold, the boundary of either a convex set or a polygon, both of which are homeomorphic to $S^1$ and, therefore, the corresponding boundary $\partial_k \freespacemappedpurging{p(j_i)}$.
\end{proof}

\begin{proof}[Proof of Theorem \ref{theorem:control_fullyactuated}]
We first focus on the proof of (the more specific) part 2 of Theorem \ref{theorem:control_fullyactuated} and follow similar patterns with the proof of \cite[Theorem 1]{vasilopoulos_pavlakos_schmeckpeper_daniilidis_koditschek_2019}. First of all, the vector field $\controlfullyactuated^\hybridmode$ is Lipschitz continuous since $\controlfullyactuatedmodel^\hybridmode(\robotpositionmodel)$ is shown to be Lipschitz continuous in \cite{Arslan_Koditschek_2018} and $\robotpositionmodel = \diffeo(\robotposition)$ is a smooth change of coordinates away from sharp corners. Therefore, the vector field $\controlfullyactuated^\hybridmode$ generates a unique continuously differentiable partial flow. To ensure completeness (i.e., absence of finite time escape through boundaries in $\freespacemapped$) we must verify that the robot never collides with any obstacle in the environment, i.e., leaves its freespace positively invariant. However, this property follows directly from the fact that the vector field $\controlfullyactuated^\hybridmode$ on $\freespacemapped$ is the pushforward of the complete vector field $\controlfullyactuatedmodel^\hybridmode$ through $(\diffeo)^{-1}$, guaranteed to insure that $\hybridfreespacemodemodel$ remain positively invariant under its flow as shown in \cite{Arslan_Koditschek_2018}, away from sharp corners on the boundary of $\freespacemapped$. Therefore, with $\hybridmode = \knownobstaclesetindex$ the terminal mode of the hybrid controller, the freespace interior $\freespacemapped$ is positively invariant under \eqref{eq:control_fullyactuated}.

Next, we focus on the critical points of \eqref{eq:control_fullyactuated}. As shown in \cite[Lemma 6]{vasilopoulos_pavlakos_schmeckpeper_daniilidis_koditschek_2019}, with $\hybridmode = \knownobstaclesetindex$ the terminal mode of the hybrid controller:
    \begin{enumerate}
        \item The set of stationary points of control law \eqref{eq:control_fullyactuated} is given as $\{ \goalposition \} \bigcup \{ (\diffeo)^{-1}(\mathbf{s}_i)\}_{i \in \knownobstaclesetdilatedmappeddiskindex} \bigcup \{\mathcal{G}_k\}_{k \in \unknownobstaclesetdilatedsemanticindex}$, where
        \begin{subequations} \label{eq:saddles}
            \begin{eqnarray}
                & \mathbf{s}_i = \diffeocenter{i} - \diffeoradius{i} \dfrac{\diffeo(\goalposition)-\diffeocenter{i}}{|| \diffeo(\goalposition)-\diffeocenter{i} ||} \label{eq:saddles_disks} \\
                & \mathcal{G}_k = \left\{ \mathbf{q} \in \freespacemapped \Big | d(\mathbf{q},\unknownobstacledilated_k)=\robotradius, \kappa(\mathbf{q}) = 1 \right\} \label{eq:saddles_convex}
            \end{eqnarray}
        \end{subequations}
        with
        \begin{equation*}
            \kappa(\mathbf{q}):=\dfrac{(\mathbf{q}-\projection{\overline{\unknownobstacledilated}_k}{\mathbf{q}})^\top(\mathbf{q}-\diffeo(\goalposition))}{||\mathbf{q}-\projection{\overline{\unknownobstacledilated}_k}{\mathbf{q}}|| \cdot ||\mathbf{q}-\diffeo(\goalposition)||}
        \end{equation*}
        \item The goal $\goalposition$ is the only locally stable equilibrium of control law \eqref{eq:control_fullyactuated} and all the other stationary points $\{ (\diffeo)^{-1}(\mathbf{s}_i)\}_{i \in \knownobstaclesetdilatedmappeddiskindex} \bigcup \{\mathcal{G}_k\}_{k \in \unknownobstaclesetdilatedsemanticindex}$, each associated with an obstacle, are nondegenerate saddles.
    \end{enumerate}

Consider the smooth Lyapunov function candidate $V^{\hybridmode}(\robotposition) = ||\diffeo(\robotposition)-\diffeo(\goalposition)||^2$. Using \eqref{eq:control_fullyactuated} and writing $\robotpositionmodel = \diffeo(\robotposition)$ and $\goalpositionmodel = \diffeo(\goalposition)$, we get
\begin{align}
\frac{dV^{\hybridmode}}{dt} = & 2(\robotpositionmodel-\goalpositionmodel)^\top\mathbf{D}_\robotposition\diffeo \, \dot{\robotposition} \nonumber \\
= & -2k(\robotpositionmodel-\goalpositionmodel)^\top \left(\robotpositionmodel - \projection{\localfreespace{\robotpositionmodel}}{\goalpositionmodel} \right) \nonumber \\
= & -2k\left(\robotpositionmodel-\projection{\localfreespace{\robotpositionmodel}}{\goalpositionmodel}+\projection{\localfreespace{\robotpositionmodel}}{\goalpositionmodel}-\goalpositionmodel \right)^\top \nonumber \\ & \left(\robotpositionmodel - \projection{\localfreespace{\robotpositionmodel}}{\goalpositionmodel} \right) \nonumber \\
= & -2k || \robotpositionmodel-\projection{\localfreespace{\robotpositionmodel}}{\goalpositionmodel} ||^2 \nonumber \\
& +2k \left( \goalpositionmodel - \projection{\localfreespace{\robotpositionmodel}}{\goalpositionmodel} \right)^\top \left(\robotpositionmodel - \projection{\localfreespace{\robotpositionmodel}}{\goalpositionmodel} \right) \nonumber \\
\leq & -2k || \robotpositionmodel-\projection{\localfreespace{\robotpositionmodel}}{\goalpositionmodel} ||^2 \leq 0
\end{align}
since $\robotpositionmodel \in \localfreespace{\robotpositionmodel}$, which implies that 
\begin{equation}
\left( \goalpositionmodel - \projection{\localfreespace{\robotpositionmodel}}{\goalpositionmodel} \right)^\top\left(\robotpositionmodel - \projection{\localfreespace{\robotpositionmodel}}{\goalpositionmodel} \right) \leq 0 \label{eq:freespaceinnerproduct}
\end{equation}
since either $\goalpositionmodel = \projection{\localfreespace{\robotpositionmodel}}{\goalpositionmodel}$, or $\goalpositionmodel$ and $\robotpositionmodel$ are separated by a hyperplane passing through $\projection{\localfreespace{\robotpositionmodel}}{\goalpositionmodel}$. Therefore, similarly to \cite{Arslan_Koditschek_2018}, using LaSalle's invariance principle we see that every trajectory starting in $\freespacemapped$ approaches the largest invariant set in $\{ \robotposition \in \freespacemapped \, | \, \dot{V}^{\hybridmode}(\robotposition)=0 \}$, i.e. the equilibrium points of \eqref{eq:control_fullyactuated}. The desired result follows directly from the fact that $\goalposition$ is the only locally stable equilibrium of our control law and the rest of the stationary points are nondegenerate saddles, whose regions of attraction have empty interior in $\freespacemapped$, as discussed above.

Next, we focus on the more general part 1 of Theorem \ref{theorem:control_fullyactuated}. Since the target now moves, we compute the time derivative of $V^{\hybridmode}$, using \eqref{eq:freespaceinnerproduct}, as
\begin{align}
\frac{dV^{\hybridmode}}{dt} = & 2(\robotpositionmodel-\goalpositionmodel)^\top\left[\mathbf{D}_\robotposition\diffeo(\robotposition) \cdot \dot{\robotposition} - \mathbf{D}_\robotposition\diffeo(\goalposition) \cdot \goalpositiondot \right] \nonumber \\
= & -2k(\robotpositionmodel-\goalpositionmodel)^\top \left(\robotpositionmodel - \projection{\localfreespace{\robotpositionmodel}}{\goalpositionmodel} \right) - 2 (\robotpositionmodel-\goalpositionmodel)^\top \goalpositionmodeldot \nonumber \\
\leq & -2k || \robotpositionmodel-\projection{\localfreespace{\robotpositionmodel}}{\goalpositionmodel} ||^2 - 2 (\robotpositionmodel-\goalpositionmodel)^\top \goalpositionmodeldot \nonumber
\end{align}
If $(\robotpositionmodel-\goalpositionmodel)^\top \goalpositionmodeldot >0$, then the desired result $\dfrac{dV^\hybridmode}{dt} \leq 0$ is immediately derived. On the other hand, if $||\goalpositionmodeldot|| \leq k \, \dfrac{|| \robotpositionmodel - \projection{\ball{\robotpositionmodel}{0.5d(\robotpositionmodel,\partial \freespacemodel)}}{\goalpositionmodel} ||^2}{||\robotpositionmodel-\goalpositionmodel||}$, then we use the Cauchy-Schwarz inequality $- 2 (\robotpositionmodel-\goalpositionmodel)^\top \goalpositionmodeldot \leq 2 || \robotpositionmodel-\goalpositionmodel || \, || \goalpositionmodeldot||$ to write
\begin{align}
    \frac{dV^{\hybridmode}}{dt} \leq & -2k || \robotpositionmodel-\projection{\localfreespace{\robotpositionmodel}}{\goalpositionmodel} ||^2 + 2 || \robotpositionmodel-\goalpositionmodel || \, || \goalpositionmodeldot|| \nonumber \\
    \leq & -2k || \robotpositionmodel-\projection{\localfreespace{\robotpositionmodel}}{\goalpositionmodel} ||^2 \nonumber \\
    & + k \, || \robotpositionmodel - \projection{\ball{\robotpositionmodel}{0.5d(\robotpositionmodel,\partial \freespacemodel)}}{\goalpositionmodel} ||^2 \label{eq:lyapunov_derivative_moving}
\end{align}
Note here that by construction of the convex local freespace in the model space $\localfreespace{\robotpositionmodel}$ as in \cite[Eqn. (25)]{Arslan_Koditschek_2018}, which guarantees that the distance of $\robotpositionmodel$ to the boundary of $\localfreespace{\robotpositionmodel}$ is $\frac{d(\robotpositionmodel,\partial \freespacemodel)}{2}$, we get that $\ball{\robotpositionmodel}{0.5 \, d(\robotpositionmodel,\partial \freespacemodel)} \subset \localfreespace{\robotpositionmodel}$.

We need to distinguish between two cases:
\begin{enumerate}
    \item If $\goalpositionmodel \in \ball{\robotpositionmodel}{0.5 \, d(\robotpositionmodel,\partial \freespacemodel)}$, then:
    \begin{equation}
        \projection{\ball{\robotpositionmodel}{0.5d(\robotpositionmodel,\partial \freespacemodel)}}{\goalpositionmodel} = \goalpositionmodel \nonumber
    \end{equation}
    and $|| \robotpositionmodel - \projection{\ball{\robotpositionmodel}{0.5d(\robotpositionmodel,\partial \freespacemodel)}}{\goalpositionmodel} || = ||\robotpositionmodel-\goalpositionmodel||$. Moreover, since $\ball{\robotpositionmodel}{0.5 \, d(\robotpositionmodel,\partial \freespacemodel)} \subset \localfreespace{\robotpositionmodel}$, $|| \robotpositionmodel-\projection{\localfreespace{\robotpositionmodel}}{\goalpositionmodel} || = ||\robotpositionmodel-\goalpositionmodel||$. From \eqref{eq:lyapunov_derivative_moving}, we now immediately get that
    \begin{equation}
        \frac{dV^{\hybridmode}}{dt} \leq -k || \robotpositionmodel-\goalpositionmodel||^2 \leq 0 \nonumber
    \end{equation}
    \item If $\goalpositionmodel \notin \ball{\robotpositionmodel}{0.5 \, d(\robotpositionmodel,\partial \freespacemodel)}$, then $|| \robotpositionmodel - \projection{\ball{\robotpositionmodel}{0.5d(\robotpositionmodel,\partial \freespacemodel)}}{\goalpositionmodel} || = \frac{d(\robotpositionmodel,\partial \freespacemodel)}{2} \leq ||\robotpositionmodel-\projection{\localfreespace{\robotpositionmodel}}{\goalpositionmodel} ||$, since $\ball{\robotpositionmodel}{0.5 \, d(\robotpositionmodel,\partial \freespacemodel)} \subset \localfreespace{\robotpositionmodel}$. The desired result $\frac{dV^{\hybridmode}}{dt} \leq 0$ is now derived from \eqref{eq:lyapunov_derivative_moving} by simple substitution.
\end{enumerate}
\end{proof}

\end{document}